\pdfoutput=1
\documentclass[twoside,11pt]{article}
\usepackage{jair, theapa, rawfonts}

\usepackage{xcolor}
\usepackage{amsmath}
\usepackage{multirow}
\usepackage{graphicx}
\usepackage{subfig}
\usepackage{amssymb}
\usepackage{algorithmicx}
\usepackage{algpseudocode}
\usepackage{algorithm}
\usepackage{appendix}

\usepackage[normalem]{ulem}
\useunder{\uline}{\ul}{}

\newcommand\blfootnote[1]{%
  \begingroup
  \renewcommand\thefootnote{}\footnote{#1}%
  \addtocounter{footnote}{-1}%
  \endgroup
}

\ShortHeadings{MADRaS : Multi Agent Driving Simulator}
{Santara \& Rudra et al.}
\firstpageno{1}

\begin{document}

\title{MADRaS : Multi Agent Driving Simulator}

\author{\name Anirban Santara$^{* \dagger}$ \email nrbnsntr@gmail.com \\
       \addr Department of Computer Science and Engineering,\\
       Indian Institute of Technology Kharagpur, Kharagpur, WB, India\\ 
      \AND
       \name Sohan Rudra$^\dagger$ \email sohanrudra@iitkgp.ac.in \\
      \addr Department of Mathematics,\\
      Indian Institute of Technology Kharagpur, Kharagpur, WB, India\\ 
      \AND
       \name Sree Aditya Buridi \email buridiaditya@iitkgp.ac.in \\
       \addr Department of Computer Science and Engineering,\\
       Indian Institute of Technology Kharagpur, Kharagpur, WB, India\\ 
      \AND
       \name Meha Kaushik \email meha.kaushik@microsoft.com  \\
       \addr Microsoft, Vancouver, Canada\\
      \AND
       \name Abhishek Naik \email abhishek.naik@ualberta.ca \\
       \addr Department of Computing Science, University of Alberta,\\ 
       Alberta, Canada\\
      \AND
       \name Bharat Kaul \email bharat.kaul@intel.com \\
       \addr Parallel Computing Lab, Intel Labs, Intel, Bengaluru, KA, India\\
      \AND
       \name Balaraman Ravindran \email ravi@cse.iitm.ac.in \\
       \addr Robert Bosch Center for Data Science and Artificial Intelligence,\\
       Indian Institute of Technology Madras, Chennai, TN, India\\
       }
\blfootnote{$^\dagger$ Authors contributed equally.\\
$^*$ Currently working at Google.\\
http://github.com/madras-simulator}

\maketitle

\pdfoutput=1
\begin{abstract}
Autonomous driving has emerged as one of the most active areas of research as it has the promise of making transportation safer and more efficient than ever before. Most real-world autonomous driving pipelines perform perception, motion planning and action in a loop. In this work we present MADRaS, an open-source multi-agent driving simulator for use in the design and evaluation of motion planning algorithms for autonomous driving. Given a start and a goal state, the task of motion planning is to solve for a sequence of position, orientation and speed values in order to navigate between the states while adhering to safety constraints. These constraints often involve the behaviors of other agents in the environment. MADRaS provides a platform for constructing a wide variety of highway and track driving scenarios where multiple driving agents can trained for motion planning tasks using reinforcement learning and other machine learning algorithms. MADRaS is built on TORCS, an open-source car-racing simulator. TORCS offers a variety of cars with different dynamic properties and driving tracks with different geometries and surface properties. MADRaS inherits these functionalities from TORCS and introduces support for multi-agent training, inter-vehicular communication, noisy observations, stochastic actions, and custom traffic cars whose behaviors can be programmed to simulate challenging traffic conditions encountered in the real world. MADRaS can be used to create driving tasks whose complexities can be tuned along eight axes in well defined steps. This makes it particularly suited for curriculum and continual learning. MADRaS is lightweight and it provides a convenient OpenAI Gym interface for independent control of each car. Apart from the primitive steering-acceleration-brake control mode of TORCS, MADRaS offers a hierarchical track-position -- speed control that can potentially be used to achieve better generalization. MADRaS uses a UDP based client server model where the simulation engine is the server and each client is a driving agent. MADRaS uses multiprocessing to run each agent as a parallel process for efficiency and integrates well with popular reinforcement learning libraries like RLLib. We show experiments on single and multi-agent reinforcement learning with and without curriculum.

\end{abstract}
\pdfoutput=1
\section{INTRODUCTION}
Inefficient driving habits of humans result in accidents, congestion and environmental pollution. These issues can be addressed efficiently if cars are able to operate autonomously. Additionally, humans lose hours of productivity in their cars towards their daily commute. These possibilities have, of late, spurred an unprecedented amount of interest towards self-driving car technology from researchers around the world.\\

Although realization of fully autonomous driving seems far flung, some specific low level tasks pertaining to driving such as adaptive cruise control, lane keep assistance and parking assistance have already been automated at a production scale in the form of Advanced Driver-Assistance Systems (ADAS) \cite{dikmen2016autonomous,minster2018system}. Safe, optimal and fast motion planning in complex, multi-modal, multi-agent, and partially observed environments is the foremost technological challenge towards achieving full autonomy. Achieving these goals tractably using traditional motion planning algorithms -- like Model Predictive Control, RRT, A$^*$, and Dijkstra -- is only possible under certain simplifying assumptions on the complexity the environment \cite{lavalle2006planning}. On the other hand, Machine Learning based approaches including Reinforcement Learning (RL) \cite{sutton2018reinforcement} and Learning from Demonstration (LfD) \cite{argall2009survey} are capable of fast, reactive control under fewer assumptions \cite{shalev2016safe,bojarski2017explaining,DBLP:journals/corr/SharifzadehCTC16,YOU20191}. However the training phase of these algorithms is often data-hungry \cite{8441797,visapp19} especially for those using highly expressive and complex models like deep neural networks. RL based methods also require online interaction with the environment that entails risk \cite{shalev2016sample,santara2017rail}. Driving simulators attempt to address these problems by rendering realistic driving conditions and traffic patterns in which agents can collect training data many times faster than real time. They also provide a sandbox environment where the agent can run into catastrophic situations while learning to drive without causing physical damage in the real world.\\

Real world driving scenarios have a high degree of variability and require the driver to optimize for multiple -- often conflicting -- objectives depending on the situation they are in. Curriculum learning \cite{bengio2009curriculum} and continual learning \cite{parisi2019continual} are two families of machine learning algorithms that are relevant in this case. Curriculum learning provides a way of learning complex skills efficiently by breaking up the problem into a hierarchy of sub-tasks and learning to accomplish them in the order of increasing complexity. Continual learning on the other hand deals with learning to accomplish new tasks without forgetting previously acquired skills. A simulator for curriculum and continual learning of autonomous driving agents should be able to create a large variety of driving scenarios with fine-grained control on their complexities.\\

\vspace{0.5cm}
Since the early days of autonomous driving research, simulators have been used in the development of different parts of the perceive-plan-act pipeline \cite{sulkowski2018search}. Most of these simulators cater to the task of perception. Back in 1989, the creators of ALVINN, Pomerleau et al. \cite{pomerleau1989alvinn}, had used a simulator to generate training images for road detection. Thanks to the recent advances in computer graphics, modern driving simulators and games like GTA-V \cite{GTA-V} can render photo-realistic driving scenes with accurate depiction of illumination, weather and other physical phenomena. They also simulate real-life sensors that can be used to collect synthetic data from these scenes to augment real-world driving datasets. Recent works \cite{chen2015deepdriving,richter2016playing,richter2017playing,ros2016synthia} have demonstrated that training perception algorithms on these augmented datasets result in better generalization in the real world that is crucial for safe and reliable autonomous driving. Most notable open-source driving simulators in this category are CARLA \cite{dosovitskiy2017carla}, Microsoft AirSim \cite{shah2018airsim}, DeepDrive.io \cite{Deepdrive} and Udacity's Self Driving Car Simulator \cite{brown2018udacity}. These simulators can, in principle, be also used for planning. However, an agent learning to face real world driving scenarios must learn to be invariant to road geometries, traffic patterns and vehicular dynamics. These simulators do not offer enough variability along these dimensions that is necessary to learn the invariances. In a typical driving scene, multiple entities (cars, buses, bikes, and pedestrians) try to achieve their objectives of getting from one place to another fast, yet safely and reliably. A simulator for such an environment should provide an easy way to create arbitrary traffic configurations. The task of negotiating in traffic is akin to finding the winning strategy in a multi-agent game \cite{dresner2008multiagent}. Hence, an autonomous driving simulator should be able to simulate different varieties of traffic and support multiple agents learning to negotiate and drive through cooperation and competition. Among the aforementioned simulators, AirSim, DeepDrive.io and Udacity provide some preset driving conditions mostly without traffic. They do not provide any straightforward way to create custom traffic or train multiple agents. CARLA does provide an API for independent control of cars that can be used for multi-agent training and creating custom traffic cars. However, most of the variability presented by CARLA is in the perceived inputs and not in the behavioral dynamics of the ego-vehicle or the traffic agents. This motivated us to develop a dedicated simulator for learning to plan in autonomous driving with a focus on learning invariances to road geometries, traffic patterns and vehicular dynamics in both single and multi-agent learning settings.\\

In this paper we present MADRaS, a \emph{M}ulti-\emph{A}gent \emph{DR}iving \emph{S}imulator for motion planning in autonomous driving and demonstrate its ability to create driving scenarios with high degrees of variability. We present results of training reinforcement learning agents to accomplish challenging tasks like driving vehicles with drastically different dynamics, maneuvering through a variety of track geometries at a high speed, navigating through a narrow road avoiding collisions with both moving and parked traffic cars and making two cars learn to cooperate and pass through a traffic bottleneck. We also demonstrate how curriculum learning can help in reducing the sample complexity of some of these tasks. Built on top of the TORCS platform \cite{wymann2000torcs}, MADRaS uses simplified physics simulation and representative graphics to reduce the computational overhead for perception and action. It allows for the addition of an unlimited number of learning and non-learning cars to a driving scene to create custom traffic configurations and train multiple agents simultaneously. Each driving agent gets a high-level object-oriented representation of the world as observation and an OpenAI gym \cite{brockman2016openai} interface for independent control. MADRaS is open source and aims to contribute to the democratization of artificial intelligence. \\

The rest of the paper is organised as follows. Section \ref{sec:mathematical_background} introduces the theoretical concepts that guide the organization of MADRaS. Section \ref{sec:madras} describes our contributions in this project in detail. Section \ref{sec:experiments} presents six experimental studies that highlight the ability of MADRaS to simulate driving tasks of high variance. Finally, Section \ref{sec:conclusion} concludes the paper with scopes of future work.

\pdfoutput=1
\vspace{0.3in}
\section{Background}
\label{sec:mathematical_background}
In this section, we introduce the concepts of Markov Decision Process (MDP), Markov Game (MG), Reinforcement Learning (RL) and Episodic Learning that comprise the foundation of MADRaS.\\

\subsection{Reinforcement Learning in Markov Games}
Markov Decision Process (MDP) is a mathematical construct that is commonly used in the artificial intelligence literature to describe an environment in which agents learn to act \cite{sutton2018reinforcement}. In a single-agent learning set-up, an MDP can be expressed as a $4$-tuple: $\mathcal{M}=\langle S, A, P, R \rangle$. It consists of a state space $S$, an action space $A$, a transition dynamics function $P:S\times A \times S\rightarrow [0, 1]$ that gives the probability distribution over next states for each action taken in a given state and a reward function $R:S\times A\rightarrow \mathbb{R}$ that qualifies the task at hand. An agent learning to act in this environment receives observations about the current state and samples actions from its policy $\pi:S \times A\rightarrow [0, 1]$ which is a conditional distribution over $A$ given a state in $S$. The reward function $R$ gives scalar feedback about these actions that indicate the agent's progress towards the goal. The agent optimizes the parameters of its policy to maximize the cumulative reward received from the environment. This form of learning through trial and error with feedback from the environment is known as Reinforcement Learning (RL).\\ 

In a multi-agent reinforcement learning set up, the environment is described as a Markov Game (MG) which is a generalization of MDP to capture the interplay of multiple agents \cite{littman1994markov,bu2008comprehensive,bowling2000analysis,zhang2019multi} . An MG is a tuple $\langle S, \{\alpha_i\}_{i=1}^n, \{A_i\}_{i=1}^n, P, \{R_i\}_{i=1}^n \rangle$. Here, $\{\alpha_i\}_{i=1}^n$ denotes a set of $n$ agents that simultaneously learn to act in an environment with state space $S$ and transition function $P$. $A_i$ and $R_i$ denote the set of actions and reward function for agent $\alpha_i$.\\ 
 

\subsection{Episodic Learning}
In episodic learning \cite{seel2011encyclopedia}, an agent's experience happens in the form of episodes. Each episode begins with the agent in one of the initial states of the environment. The state of the environment changes in response to the agent's actions and the episode ends when the environment sends a \emph{done} signal to the agent. In a general multi-agent learning setting, the environment may send a done signal to each agent separately at different time steps resulting in different episode lengths for each agent. When the episodes of all the agents end, the environment resets itself to one of its initial states and starts new episodes for each agent.
\pdfoutput=1
\section{MADRaS Simulator}
\label{sec:madras}

\begin{table}[]
\centering
\caption{Comparison of Gym TORCS \cite{yoshida2016gym} with MADRaS}
\label{table:torcs-vs-madras}
\begin{tabular}{|l|c|c|}
\hline
\textbf{Feature} & \textbf{Gym TORCS} & \textbf{MADRaS} \\ \hline
scr-server architecture & $\checkmark$ & $\checkmark$ \\ \hline
observation noise & $\times$ & $\checkmark$ \\ \hline
stochastic outcomes of actions & $\times$ & $\checkmark$ \\ \hline
parallel rollout support & $\times$ & $\checkmark$ \\ \hline
multi-agent training & $\times$ & $\checkmark$ \\ \hline
inter-vehicular communication & $\times$ & $\checkmark$ \\ \hline
custom traffic cars & $\times$ & $\checkmark$ \\ \hline
domain randomization & $\times$ & $\checkmark$ \\ \hline
centralized configuration & $\times$ & $\checkmark$ \\ \hline
modular reward and done functions & $\times$ & $\checkmark$ \\ \hline
hierarchical action space & $\times$ & $\checkmark$ \\ \hline
\end{tabular}
\end{table}

In this section we describe the structure and organization of the MADRaS simulator which constitutes the main contribution of this paper. The current version of MADRaS is focused on track driving. Track driving is traditionally used in the automotive world to benchmark driver skill and car agility. We first present a brief overview of the TORCS simulator and associated prior works that MADRaS builds upon. Then we describe the new features that we develop in this project and present a thorough empirical analysis of their relevance in the context of planning in autonomous driving. 

\begin{figure}
    \centering
    \includegraphics[width=0.8\textwidth]{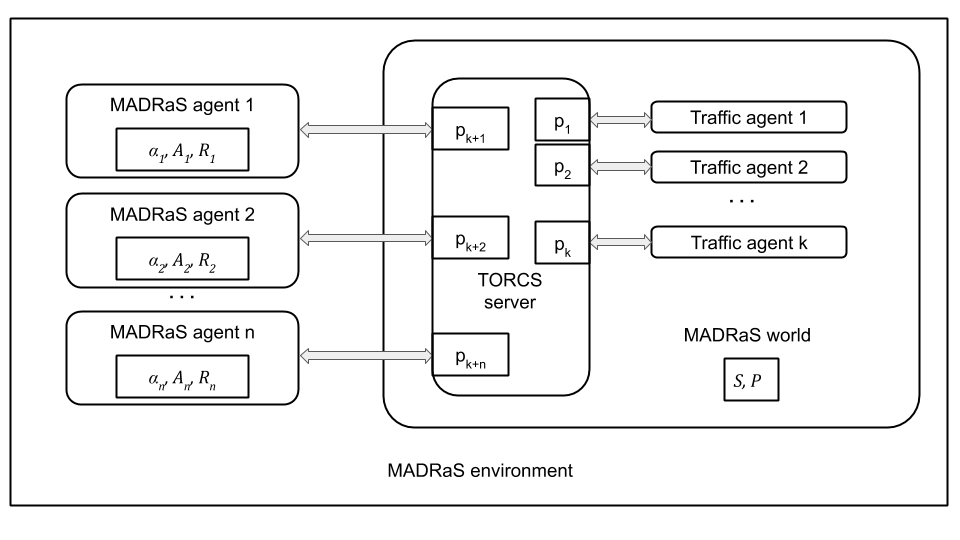}
    \caption{Architecture of the MADRaS simulation environment. Each double headed arrow indicates one UDP communication channel between the TORCS server and one of the clients (traffic or MADRaS agents). The server listens to the $i^{th}$ client through a dedicated port denoted by p$_i$ in the figure. MADRaS assigns these ports in order, first to the traffic agents and then to the learning agents. The Markov Game terms are also marked in their respective places of definition in the figure.}
    \label{fig:madras-arch}
\end{figure}{}

\subsection{TORCS Simulator}

MADRaS is based on TORCS which stands for The Open Racing Car Simulator \cite{wymann2000torcs}. It is capable of simulating the essential elements of vehicular dynamics such as mass, rotational inertia, collision, mechanics of suspensions, links and differentials, friction and aerodynamics. Physics simulation is simplified and is carried out through Euler integration of differential equations at a temporal discretization level of $0.002$ seconds. The rendering pipeline is lightweight and based on OpenGL \cite{neider1993opengl} that can be turned off for faster training. TORCS offers a large variety of tracks and cars as free assets that we discuss later in this section. It also provides a number of programmed robot cars with different levels of performance that can be used to benchmark the performance of human players and software driving agents. TORCS was built with the goal of developing Artificial Intelligence for vehicular control and has been used extensively by the machine learning community ever since its inception 
\cite{li2017infogail,lillicrap2015continuous,loiacono2010learning,koutnik2013evolving,koutnik2014evolving,onieva2010overtaking}.\\

\subsection{SCR Server-Client Architecture}

The Simulated Car Racing (SCR) Championship \cite{loiacono20102009} is an annual car-racing competition where participants submit controllers for racing in the TORCS environment. It provides a software patch for TORCS known as \texttt{scr\_server} \cite{loiacono2013simulated}. It sets up a UDP based client-server architecture in which the competing cars can operate independent of one another. The server runs the TORCS simulator. Each client represents a car that runs as a separate process and communicates with the server through a dedicated UDP port. The patch also provides a layer of abstraction over TORCS in which each car has access to an egocentric view of the environment and not the entire game state. The server polls actions from the clients and updates the game-state every $0.02$ seconds of simulated time. The official build of TORCS supports up to $10$ SCR clients at a time but with modifications like in \cite{kaushik2018overtaking} the number of clients can be increased arbitrarily.\\ 

\subsection{GymTORCS Environment}

GymTORCS \cite{yoshida2016gym} is an OpenAI Gym \cite{brockman2016openai} wrapper for SCR cars built for use in Reinforcement Learning experiments. It uses a custom library called \emph{Snake Oil} to create a client for communicating with the TORCS server through the \texttt{scr\_server} interface. Snake Oil also provides plug-ins for automatic-transmission, traction control and throttle control which can be used to provide different control modes to the driving agent. GymTORCS is popular in the reinforcement learning community for experiments on driving tasks \cite{kaushik2018overtaking,liu2017learning,de2018integrating,dossa2019human}. MADRaS builds on GymTORCS by increasing its stability and ease of use and adding features like multi-agent training and custom traffic cars.\\

\subsection{MADRaS: Multi-Agent Driving Simulator}

Having described TORCS and associated prior works that form the foundation of MADRaS, we now present our contributions in this project. As GymTORCS is pre-dominantly designed for single-agent training, the environment is inherently structured as an MDP. This restricts its usage for multi-agent training. MADRaS is GymTORCS restructured as an MG with some added functionalities (see Table \ref{table:torcs-vs-madras}). Figure \ref{fig:madras-arch} describes the architecture of MADRaS. \emph{MADRaS Environment} consists of a \emph{MADRaS World} and a given number of \emph{MADRaS Agents} $(\{\alpha_i\}_i)$. \emph{MADRaS World} consists of a TORCS server and a given number of traffic agents each of which executes an independently configured behavior. The state space $(S)$ and the transition dynamics $(P)$ of the MG are defined by the \emph{MADRaS World}. Each \emph{MADRaS Agent} $\alpha_i$ runs as an SCR Client with a modified Snake Oil interface that has its own action space $A_i$ and reward function $R_i$ which are independent of the action spaces and reward functions of the other agents. Unlike GymTORCS, \emph{MADRaS Agents} can not reset the TORCS server. This allows for multiple agents to complete their episodes independently. \emph{MADRaS Environment} resets its \emph{MADRaS World} and in turn its TORCS server when all the agents have terminated their episodes. MADRaS also provides a number of ways to configure the initial state of the environment for the task at hand. The initial distance from the start line and position with respect to the track edges can be specified individually for both the learning cars as well as the traffic agents. Thus MADRaS harnesses the full potential of the SCR server-client architecture and enables multi-agent training. We describe the salient features of MADRaS in the remaining part of this section.

\subsubsection{Traffic Agents}
MADRaS introduces support for adding non-learning traffic agents in the environment that execute a pre-defined behaviour. These are different from the robot cars that come bundled with TORCS for benchmarking racing agents. MADRaS provides a base class that can be used as template to create traffic cars with interesting behavioral patterns and some sample traffic classes as free assets (see Table \ref{table:traffic-agents}). The base class also comes equipped with methods to prevent collision and going out of track. Each traffic agent runs as a parallel process independent of the learning agent and has an SCR client that talks to the TORCS server through a dedicated port. MADRaS takes care of the configuration and assignment of a requisite number of server ports for connecting all the learning and traffic agents properly at the start of each episode. The number and behavior of traffic agents can be varied between episodes.\\ 

\begin{table*}[t]
\centering
\caption{Sample traffic agents in MADRaS.}
\label{table:traffic-agents}
\begin{tabular}{|l|p{3.5in}|}
\hline
\multicolumn{1}{|c|}{\textbf{Name}} & \multicolumn{1}{c|}{\textbf{Behaviour}} \\ \hline
\texttt{ConstVelTrafficAgent} & Drives at a given speed at a given lane position. \\ \hline
\texttt{SinusoidalSpeedAgent} & Varies the speed sinusoidally while driving at a given lane position. \\ \hline
\texttt{RandomLaneSwitchAgent} & Agent switches lanes randomly while driving. \\ \hline
\texttt{DriveAndParkAgent} & Agent drives to a given distance and track-position and parks itself. \\ \hline
\texttt{ParkedAgent} & Agent remains parked at a given distance and track-position throughout. \\ \hline
\texttt{RandomStoppingAgent} & Agent halts randomly while driving. \\ \hline
\end{tabular}
\end{table*}

\begin{figure*}[!t]
    \centering
    \includegraphics[width=0.85\textwidth]{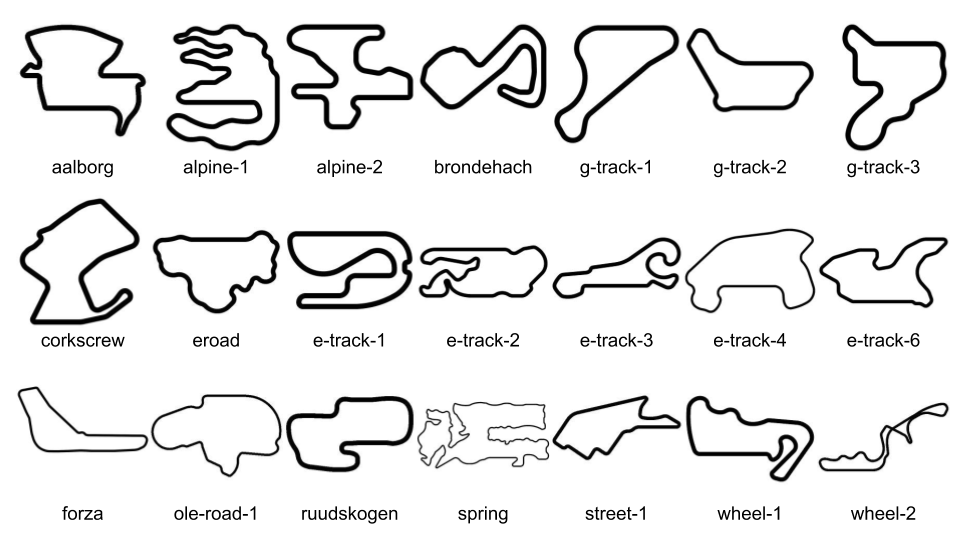}
    \caption{Schematic diagrams of road tracks in TORCS \cite{wymann2000torcs}.}
    \label{fig:torcs-roads}
\end{figure*}

\subsubsection{Tracks}
One of the major advantages of TORCS as the platform of choice for building MADRaS is the availability of a large number of tracks with different geometric (see Figure \ref{fig:torcs-roads}) and surface properties. At the time of writing this paper, TORCS offers $9$ oval, $21$ road, and $8$ dirt tracks. MADRaS inherits these free assets from the TORCS project. A limitation of GymTORCS is that a track has to be chosen at the beginning of a training experiment and it remains fixed throughout. This often causes the agent to memorize the track resulting in poor generalization. MADRaS ameliorates this by introducing an option to select a track at the beginning of each episode. Thus the agent can be exposed to multiple tracks during training.\\ 

\subsubsection{Car Models}
TORCS provides $42$ car models with a wide range of dynamic properties. However, GymTORCS only supports a single default car type named \texttt{car1-trb1}. MADRaS is capable of changing cars at the beginning of each training episode. Thus it makes it possible to train an agent to drive cars with drastically different dynamic properties. Also, the learning and traffic agents can be assigned different car types for visual distinction.\\

\subsubsection{Modular Configuration}
As Reinforcement Learning (RL) is one of the most powerful and actively researched approaches for robot motion planning, MADRaS has some features tailor-made for that purpose. The exercise of tuning an RL algorithm for a given task usually involves tweaking the reward function and episode termination (``done") criteria. It is important to keep accurate track of these parameters across experiments to be able to arrive at the optimal training configuration. GymTORCS has particularly poor configurability as it requires the user to make changes in the Python source code which are difficult to keep track of. The entire MADRaS environment including the reward and done functions are configurable through a single file named \texttt{madras\_config.yml}. A copy of this configuration file can be saved in the training directory for effortless tracking across experiments. Please refer to Appendix A for some commonly used configuration variables.\\

The reward and done functions are usually composed of multiple parts that try to capture events like arrival at the goal state, crashes and damages. Modularity of these definitions in code is essential for fast iteration. MADRaS provides \texttt{MadrasReward} and \texttt{MadrasDone} base classes as templates for defining the components of the reward and done functions. Specifying a reward or done function in MADRaS is as simple as listing the names 
of their components in the configuration file. Each MADRaS Agent comes with a \texttt{reward\_handler} and a \texttt{done\_handler} that organize the listed components and set up the corresponding functions. This modular architecture makes it easy to define new reward and done functions and plug them in and out of experiments easily.\\

\subsubsection{Observation Space}
The Snake Oil library of GymTORCS provides a parser for the state information returned by the TORCS server. These state variables include odometry, range data, obstacle detection, engine statistics and metadata regarding the position of the ego vehicle relative to the other cars on the road. Such a high-level representation of the world is common in practical autonomous driving pipelines \cite{bansal2018chauffeurnet} as it helps in decoupling the perception and planning modules allowing them to be improved independently and also reduces the sample complexity of machine learning based planning algorithms \cite{shalev2016sample}. Raw visual inputs in the form of a stream of images are also available. For a full list of state variables please refer to the Simulated Car Race Championship paper \cite{loiacono2013simulated}. 
The observation vector of a MADRaS agent is composed of a selection of these normalized state variables. For modularity and ease of configuration, MADRaS provides an \texttt{observation\_handler} class that can toggle between different sets of observed variables. The observations can optionally be made noisy to simulate a partially observed driving scenario.\\

\subsubsection{Action Space}
The Snake Oil library allows GymTORCS agents to control cars via steering, acceleration and brake commands. MADRaS inherits this primitive control mode and also adds a hierarchical track-position -- speed control mode. In track-position -- speed control mode, a MADRaS agent produces its desired position with respect to the left and right edges of the track and its desired speed. A PID controller takes these non-primitive actions (\emph{desires}) as inputs and calculates a sequence of steering, acceleration and brake commands in feedback mode over a number of time steps denoted by \texttt{PID\_latency}. The \texttt{PID\_latency} controls the relative time scales of the higher and lower level action spaces. The following is the expression of a PID controller for control variable $u$.

\begin{equation}
    u(t) = K_p e(t) + K_i \int_{0}^{t} e(t')dt' + K_d \frac {de(t)}{dt} 
\end{equation}

\noindent $K_p$, $K_i$ and $K_d$ are the constants for the proportional, integral and derivative terms respectively. The track-position -- speed action space is inspired by \cite{shalev2016safe}, where the authors note that training an RL agent to generate high-level desires while relegating the low-level implementation of the desires to an analytical controller like PID significantly reduces real world risk and increases the explainability of the agent's behavior. High level actions also have been reported to show better generalizability across vehicular platforms \cite{behere2016functional}. All actions are normalized between $-1$ and $1$ for ease of optimization of neural network policies. The outcomes of the agent's actions can optionally be made stochastic. MADRaS implements this stochasticity by adding zero-mean Gaussian noise to actions before sending them to the TORCS server.\\

\subsubsection{Inter-vehicular Communication}
The most salient feature of MADRaS is its support for multi-agent training. The success of multi-agent learning is contingent on the ability of the agents to communicate among themselves and plan actions taking into account the states and actions of the other agents \cite{lowe2017multi}. MADRaS provides a highly flexible framework for inter-vehicular communication through a communication buffer and an agent mapping function. The agent mapping function allows the user to specify a list of variables that the $i^{th}$ agent wants to observe from the $j^{th}$ agent. The communication buffer records these shared variables from the step $t-1$ and makes them a part of the agents' observation vectors at step $t$.\\

\subsubsection{Curriculum Design for Driving Agents}
MADRaS has been designed to provide a playground for reinforcement learning agents to learn to drive any car on any track in any kind of traffic within the TORCS environment. In order to construct a driving problem of high variance, MADRaS can present an agent with a different car to drive in a different track with a different number of traffic cars of different behaviors chosen randomly or in a given order in every training episode. MADRaS can also produce additional stochasticity by making the outcome of an action probabilistic. Training deep neural network policies in high variance environments poses a highly non-convex problem that is difficult to optimize. Curriculum learning \cite{bengio2009curriculum} has been shown to be effective in reducing the sample complexity in such problems. Curriculum learning involves training an agent on a sequence of tasks of increasing complexity. MADRaS is designed with curriculum learning in mind. The complexity of the driving task in MADRaS can be systematically increased in well defined steps along the following eight dimensions:
\begin{enumerate}
    \item Number of learning agents.
    \item Number of cars to be presented to the agent to drive.
    \item Number of tracks to be presented to the agent to drive.
    \item Number of traffic agents.
    \item Level of obstructive behavior from the traffic agents.
    \item Target speed of the learning agent(s).
    \item Degree of stochasticity to action-outcomes.
    \item Presence of noise in observations.
\end{enumerate}

\noindent In the following section we present a set of experiments to highlight the key features of MADRaS.
\section{Experiments}
\label{sec:experiments}

\begin{table}[!t]
\centering
\caption{Parameters of the PID controller used in our experiments.}
\label{table:PID-params}
\begin{tabular}{l|c|c|c|}
\cline{2-4}
 & \textbf{$K_p$} & \textbf{$K_i$} & \textbf{$K_d$} \\ \hline
\multicolumn{1}{|l|}{\textbf{acceleration PID}} & 10.5 & 0.05 & 2.8 \\ \hline
\multicolumn{1}{|l|}{\textbf{steering PID}} & 5.1 & 0.001 & 0.000001 \\ \hline
\end{tabular}
\end{table}

In this section we present the results of six experiments on single and multi-agent RL for learning to drive in MADRaS. The purpose of these experiments is to highlight the features of MADRaS that were discussed in the previous section as an improvement over GymTORCS.

\subsection{Experimental Setup}
We demonstrate how MADRaS can be used to create a wide variety of driving tasks that can be addressed by RL. We use the Proximal Policy Optimization (PPO) algorithm \cite{schulman2017proximal} for RL in all our experiments. PPO is a trust-region based local policy optimization algorithm that has been shown to be very effective in learning policies for continuous control tasks \cite{andrychowicz2020learning}. We save the comparison of different RL algorithms on MADRaS tasks for a future paper in the interest of brevity. All the performance statistics presented in this section are estimated over at least $100$ episodes. All experiments with the track-position -- speed action space have a \texttt{PID\_latency} of $5$ time steps. The reward functions of the RL agents are defined as weighted sums of reward $(r)$ and penalty $(p)$ components with weights $w_r$ and $w_p$, respectively:

\begin{equation}
    agent\_reward = \sum_{r\in rewards} w_r r - \sum_{p \in penalties} w_p p
\end{equation}

\noindent Some general purpose reward and penalty components that are used in all the experiments are as follows:\\

\textbf{Progress Reward:} Progress Reward rewards the agent for making a finite progress at every time step. We calculate progress relative to a target speed. We reward the agent proportional to its speed until it reaches the target speed. If the speed goes beyond the target speed, we do not give the agent any extra reward. This way we prevent the agent from maximizing its cumulative rewards by running fast and crashing rather than finishing the race. Let $d(t)$ be the distance (in meters) covered by the agent in the $t^{th}$ time step and $s_{target}$ denote the target speed in meters per step. Progress reward is given by:

        \begin{equation}
            progress\_reward(t) = \min\left(1, \frac{d(t)}{s_{target}}\right)
        \end{equation}
    
\textbf{Average Speed Reward:} Average Speed Reward rewards the agent for maintaining a high average speed only if it manages to complete a full lap of the track. Suppose the average speed of the agent for a lap is $s_{avg}$. Average Speed Reward is calculated as:

        \begin{equation}
            average\_reward = \frac{s_{avg}}{s_{target}}
        \end{equation}
The Average Speed Reward is also scaled (but not capped) relative to the target speed $s_{target}$ of the agent. \\

\textbf{Angular Acceleration Penalty:} This penalty is meant to discourage the agent from making frequent unnecessary side-wise movements while running down a track. We calculate a numerical approximation of angular acceleration from the the past $3$ recorded values of the \emph{angle} between the car's direction and the  direction of the track axis. We scale the penalty with respect to a reference $\alpha_{reference}$. Let $a_{t-2}, a_{t-1}, a_t$ be three consecutive angles of the agent. We calculate Angular Acceleration Penalty as:
    
        \begin{equation}
            angular\_accleration\_penalty(t) = \frac{\left| a_t+a_{t-2}-2a_{t-1} \right|}{\alpha_{reference}}
        \end{equation}
    
We set $\alpha_{reference}$ to $2.0$ in all our experiments.\\

\textbf{Turn Backward Penalty:} A fixed penalty of $-1$ if the car turns backwards.\\

\textbf{Collision Penalty:} A fixed penalty of $-1$ if the car collides with obstacles or other cars and incurs a damage.\\

\noindent Apart from these we also use task specific rewards that we define separately in each experiment.\\ 

We terminate an episode if one of the following events happen:
\begin{itemize}
    \item car turns backwards,
    \item car goes out of track,
    \item car collides with an obstacle,
    \item agent fails to complete its task within the maximum allowable duration of an episode,
    \item agent successfully completes the task at hand.\\
\end{itemize}

Unless otherwise stated, we set the learning rate to $5\times 10^{-5}$. The policy and value functions are modelled using fully connected neural networks with $2$ hidden layers and $256$ $tanh$--units in each layer.  We use the PPO implementation of RLLib \cite{liang2017rllib} for all our experiments for its stability and support for multi-agent training. The PID parameters used for track-position -- speed control are given in Table \ref{table:PID-params}. Although ideally these parameters must be tuned for each car and for each speed range, we use the same set of parameters (originally tuned for medium-low speeds of \texttt{car1-trb1}) everywhere to check if it is possible to teach RL agents to be robust to imperfections in the low level controller.\\

In the remaining part of this section, we describe our experiments and discuss the major observations\footnote{Accompanying video: https://youtu.be/io5mP0HUytY}.\\

\subsection*{Experiment 1: Generalization across tracks with higher level actions}
\label{sec:gen-tracks}

\begin{table}[!t]
\centering
\caption{RL training criteria for Experiments 1-3. Please refer to \cite{loiacono2013simulated} for details on the observed variables.}
\label{table:experiment-1-training-criteria}
\begin{tabular}{|l|l|c|}
\hline
\multirow{6}{*}{\textbf{Reward function}} & \textbf{Reward Function Component} & \textbf{Weightage} \\ \cline{2-3} 
 & Progress Reward & 1.0 \\ \cline{2-3} 
 & Average Speed Reward & 1.0 \\ \cline{2-3} 
 & Collision Penalty & 10.0 \\ \cline{2-3} 
 & Turn Backward Penalty & 10.0 \\ \cline{2-3} 
 & Angular Acceleration Penalty & 5.0 \\ \hline\hline
\textbf{Observed variables} & \multicolumn{2}{p{3in}|}{angle, track, trackPos, speedX, speedY, speedZ} \\ \hline\hline
\textbf{Done criteria} & \multicolumn{2}{p{3in}|}{One Lap Completed, Time Out, Collision, Turn Backward, Out of Track} \\ \hline
\end{tabular}
\end{table}

In our first experiment, we compare two RL agents, one having the high-level track-position -- speed (T-S) control mode and the other having the low-level steer -- acceleration -- brake (S-A-B) control mode, on their ability to generalize across multiple driving tracks in MADRaS. We train the agents to drive \texttt{car1-stock1} in the \texttt{Alpine-1} track and evaluate them on the other road tracks. Table \ref{table:experiment-1-training-criteria} lists the observed variables and the components of the reward and done functions. We set the maximum duration of an episode at $15000$ time steps and the target speed at $100$ km/hour. We evaluate the agents in terms of the average fraction of lap covered in an episode, average speed and successful lap completion rate.\\

Table \ref{table:gen-road-tracks} presents the results of this experiment. We see that the agent with high-level track-position -- speed (T-S) control generalizes significantly better than the one with low-level steer -- acceleration -- brake (S-A-B) control as given by higher average scores. The low-level S-A-B control mode gives the agent tighter control of the car that can be exploited to perform maneuvers very specific to the training track in order to navigate the twists and turns while maintaining a high average speed (see the accompanying video). This results in the agent overfitting to the training track and it fails to make any significant progress in some of the test tracks.
Implementing a desired track-position and speed may require different sequences of low-level actions in different tracks. Relegating the low-level control to a PID controller gives the T-S agent better generalization to track-geometries than the S-A-B agent.\\

\subsection*{Experiment 2: Generalization across vehicular dynamics through random car selection}
\label{sec:gen-cars}

\begin{figure}
    \centering
    \subfloat[]{{\includegraphics[width=0.7\textwidth]{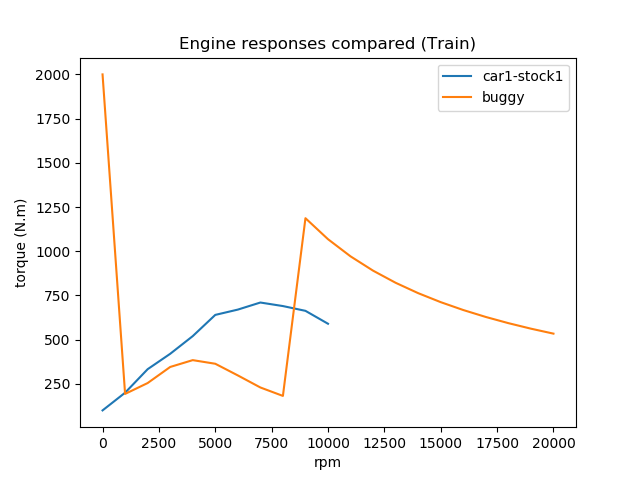} }}%
    \qquad
    \subfloat[]{{\includegraphics[width=0.7\textwidth]{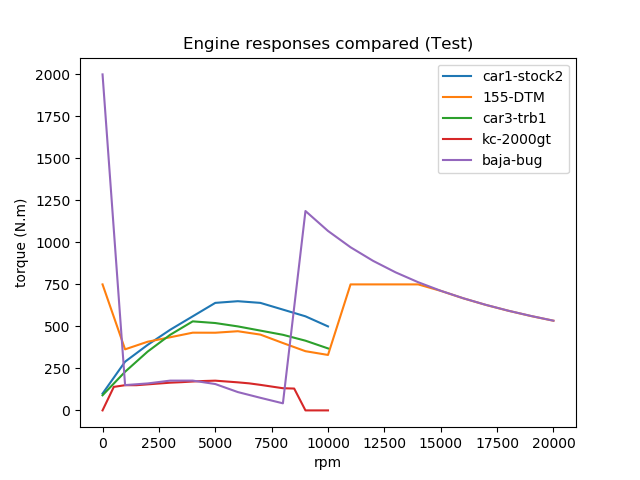} }}%
    \caption{Variation of torque with engine RPM of cars studied in Experiment 2. (a) Torque-vs-RPM of the cars that we present our agent to drive during training with equal probability. (b) Torque-vs-RPM of the cars that we test our agent on. }%
    \label{fig:throttle-responses}%
\end{figure}

In our second experiment, we leverage the ability of MADRaS to change the agent's car at the beginning of each episode to train a driving policy that generalizes to multiple cars with significantly different vehicular dynamics. Table \ref{table:car-physical-stats} gives some physical parameters of the cars used in this experiment that characterize their handling and dynamics. Heavier cars with a low centre of gravity e.g. \texttt{car1-stock1}, \texttt{car3-trb1} and \texttt{car1-stock2} are more stable and handle better with less body-roll around tight corners. The variation of torque with the RPM (Rotations Per Minute) of a car's engine plays a crucial role in deciding its dynamics. The torque produced by an engine decides how fast the car can accelerate. Torque is usually a strong function of engine RPM. While running at a given RPM, a car can accelerate faster if its engine can produce higher torque at that RPM. Figure \ref{fig:throttle-responses} gives the torque-RPM curves for the cars used in this experiment. The cars fall in two broad categories in terms of the overall shape of this curve. Cars with a ``$\cup$"-shaped curve e.g. \texttt{buggy}, \texttt{baja-bug} and \texttt{155-DTM} have high torque at low $(<1000)$ and high $(>10000)$ RPM and significantly lower values in the middle. The other category of cars e.g. \texttt{car1-stock1}, \texttt{car3-trb1} and \texttt{car1-stock2} have a ``hat" ($\cap$)-shaped curve with low torque at low and high RPM and high values in the middle. When the agent needs high torque to accelerate from a standstill, speed up or climb uphill, it needs to take the engine RPM to the high-torque zone with a suitable sequence of accelerator inputs. The high-torque zones of the aforementioned categories of cars are roughly opposite to one another. This makes it challenging for a driving agent to generalize to both kinds of cars.\\

We choose the \texttt{Alpine-1} track for this experiment. The \texttt{Alpine-1} track is one of the hardest road tracks of MADRaS with sharp left and right turns and a few stretches of slippery road. We set the maximum duration of an episode to $20000$ time steps and the target speed to $100$ km/hour. We evaluate the agent in terms of average fraction of the lap covered per episode and average speed.\\

\vspace{0.25cm}
First, we train two PPO agents to drive \texttt{car1-stock1} and \texttt{buggy} using the S-A-B control mode. We evaluate them on five test cars of different dynamic properties. Table \ref{table:poor-gen-throttle} presents the results. We see that an agent trained on a car of one torque-RPM category has difficulty generalizing to the cars of the other category. 
In our next step, with a view to aiding in generalization through domain randomization, we leverage the ability of MADRaS to randomly switch cars between episodes and present \texttt{car1-stock1} and \texttt{buggy} to the same agent with equal probability. We observe that this training strategy brings significant improvement both in terms of average fraction of lap covered in an episode and average speed. 
\\ 

\begin{table}[]
\centering
\caption{Some physical properties of the cars used in Experiment 2 that play an important role in determining their vehicular dynamics. ``RWD" and ``4WD" stand for ``Rear Wheel Drive" and ``Four Wheel Drive", respectively.}
\label{table:car-physical-stats}
\begin{tabular}{|l|c|c|c|}
\hline
\multicolumn{1}{|c|}{\textbf{Car Name}} & \textbf{Drive Type} & \textbf{Mass (Kg)} & \textbf{Height of CG (m)} \\ \hline
car1-stock1 & RWD & 1550.0 & 0.3 \\ \hline
car1-stock2 & RWD & 1550.0 & 0.3 \\ \hline
155-DTM & 4WD & 1100.0 & 0.2 \\ \hline
car3-trb1 & RWD & 1150.0 & 0.2 \\ \hline
kc-2000gt & RWD & 1200.0 & 0.25 \\ \hline
buggy & RWD & 650.0 & 0.45 \\ \hline
baja-bug & RWD & 600.0 & 0.35 \\ \hline
\end{tabular}
\end{table}

\begin{table*}[]
\centering
\caption{Generalization of an agent trained on \texttt{Alpine-1} to other road tracks (Experiment 1).}
\label{table:gen-road-tracks}
\scriptsize
\begin{tabular}{lc|c|c|c|c|c|c|}
\cline{3-8}
\multicolumn{1}{c}{\textbf{}} & \textbf{} & \multicolumn{2}{p{0.8in}|}{\textbf{Avg. fraction of lap covered}} & \multicolumn{2}{c|}{\textbf{Avg. Speed}} & \multicolumn{2}{p{0.8in}|}{\textbf{Lap completion rate}} \\ \cline{2-8} 
\multicolumn{1}{c|}{} & \textbf{Action Space} & \textbf{S-A-B} & \textbf{T-S} & \textbf{S-A-B} & \textbf{T-S} & \textbf{S-A-B} & \textbf{T-S} \\ \hline
\multicolumn{1}{|l|}{\textbf{Training Track}} & \textit{alpine-1} & \textit{0.75} & \textit{0.73} & \textit{91.89} & \textit{83.32} & \textit{0.68} & \textit{0.58} \\ \cline{1-8}
\multicolumn{8}{l}{}   \\ \cline{1-8}
\multicolumn{1}{|l|}{\multirow{20}{*}{\textbf{Test Tracks}}} & aalborg & 0.001 & 0.11 & 0.10 & 59.39 & 0.0 & 0.0 \\ \cline{2-8} 
\multicolumn{1}{|l|}{} & alpine-2 & 0.38 & 0.31 & 89.95 & 72.64 & 0.04 & 0.0 \\ \cline{2-8} 
\multicolumn{1}{|l|}{} & brondehach & 0.001 & 0.72 & 0.1 & 81.01 & 0.0 & 0.3 \\ \cline{2-8} 
\multicolumn{1}{|l|}{} & g-track-1 & 0.001 & 0.98 & 0.06 & 79.42 & 0.0 & 0.91 \\ \cline{2-8} 
\multicolumn{1}{|l|}{} & g-track-2 & 0.002 & 0.97 & 0.11 & 75.99 & 0.0 & 0.95 \\ \cline{2-8} 
\multicolumn{1}{|l|}{} & g-track-3 & 0.001 & 0.84 & 0.09 & 79.90 & 0.0 & 0.44 \\ \cline{2-8} 
\multicolumn{1}{|l|}{} & corkscrew & 0.0008 & 0.64 & 0.06 & 81.39 & 0.0 & 0.0 \\ \cline{2-8} 
\multicolumn{1}{|l|}{} & e-road & 0.001 & 0.94 & 0.11 & 85.63 & 0.0 & 0.88 \\ \cline{2-8} 
\multicolumn{1}{|l|}{} & e-track-2 & 0.07 & 0.39 & 8.38 & 75.21 & 0.0 & 0.0 \\ \cline{2-8} 
\multicolumn{1}{|l|}{} & e-track-3 & 0.31 & 0.68 & 25.88 & 77.96 & 0.03 & 0.57 \\ \cline{2-8} 
\multicolumn{1}{|l|}{} & e-track-4 & 0.0005 & 0.95 & 0.08 & 78.41 & 0.0 & 0.85 \\ \cline{2-8} 
\multicolumn{1}{|l|}{} & e-track-6 & 0.0009 & 0.83 & 0.09 & 80.65 & 0.0 & 0.58 \\ \cline{2-8} 
\multicolumn{1}{|l|}{} & forza & 0.001 & 0.79 & 0.08 & 71.63 & 0.0 & 0.70 \\ \cline{2-8} 
\multicolumn{1}{|l|}{} & ole-road-1 & 0.29 & 0.40 & 101.22 & 78.06 & 0.0 & 0.11 \\ \cline{2-8} 
\multicolumn{1}{|l|}{} & ruudskogen & 0.97 & 0.97 & 100.87 & 81.15 & 0.95 & 0.93 \\ \cline{2-8} 
\multicolumn{1}{|l|}{} & street-1 & 0.03 & 0.87 & 1.76 & 74.67 & 0.0 & 0.67 \\ \cline{2-8} 
\multicolumn{1}{|l|}{} & wheel-1 & 0.0009 & 0.95 & 0.09 & 78.08 & 0.0 & 0.76 \\ \cline{2-8} 
\multicolumn{1}{|l|}{} & wheel-2 & 0.36 & 0.81 & 81.69 & 81.51 & 0.0 & 0.64 \\ \cline{2-8} 
\multicolumn{1}{|l|}{} & spring & 0.14 & 0.29 & 104.76 & 82.55 & 0.0 & 0.0 \\ \hline
\multicolumn{1}{|l}{} & \textbf{Average Scores (Test)} & \textbf{0.14} & \textbf{0.71} & \textbf{27.12} & \textbf{77.64} & \textbf{0.04} & \textbf{0.49} \\ \hline
\end{tabular}
\end{table*}

\begin{table*}[]
\centering
\scriptsize
\caption{Generalization of PPO policies across vehicles with different dynamics (Experiment 2). ``random" refers to the setting in which the agent is presented with both \texttt{car1-stock1} and \texttt{buggy}, each with a probability of 0.5 during training.}
\label{table:poor-gen-throttle}
\begin{tabular}{ll|c|c|c|c|c|c|}
\cline{3-8}
 &  & \multicolumn{3}{c|}{\textbf{Avg. Fraction of Track Covered}} & \multicolumn{3}{c|}{\textbf{Avg. Speed (km/h)}} \\ \cline{2-8} 
\multicolumn{1}{l|}{\textbf{}} & \textbf{Training Car} & \textbf{car1-stock1} & \textbf{buggy} & \textbf{random} & \textbf{car1-stock1} & \textbf{buggy} & \textbf{random} \\ \hline
\multicolumn{1}{|l|}{\multirow{5}{*}{\rotatebox[origin=c]{90}{\textbf{Test Cars}}}} & \textbf{155-DTM} & 0.37 & 0.01 & 0.37 & 104.22 & 2.14 & 99.78 \\ \cline{2-8} 
\multicolumn{1}{|l|}{} & \textbf{car3-trb1} & 0.002 & 0.003 & 0.62 & 0.12 & 0.25 & 58.95 \\ \cline{2-8} 
\multicolumn{1}{|l|}{} & \textbf{kc-2000gt} & 0.77 & 0.003 & 0.30 & 80.44 & 0.24 & 22.02 \\ \cline{2-8} 
\multicolumn{1}{|l|}{} & \textbf{car1-stock2} & 0.001 & 0.003 & 0.54 & 0.09 & 0.23 & 50.23 \\ \cline{2-8} 
\multicolumn{1}{|l|}{} & \textbf{baja-bug} & 0.35 & 0.04 & 0.55 & 59.45 & 38.40 & 54.91 \\ \hline
\multicolumn{2}{|r|}{\textbf{Average Scores}} & 0.30 & 0.01 & \textbf{0.48} & 48.86 & 8.25 & \textbf{57.18} \\ \hline
\end{tabular}
\end{table*}

\subsection*{Experiment 3: Curriculum learning for driving in \texttt{Spring} track}
\label{sec:curriculum-spring}
\begin{figure*}[!t]
    \centering
    \includegraphics[width=\textwidth]{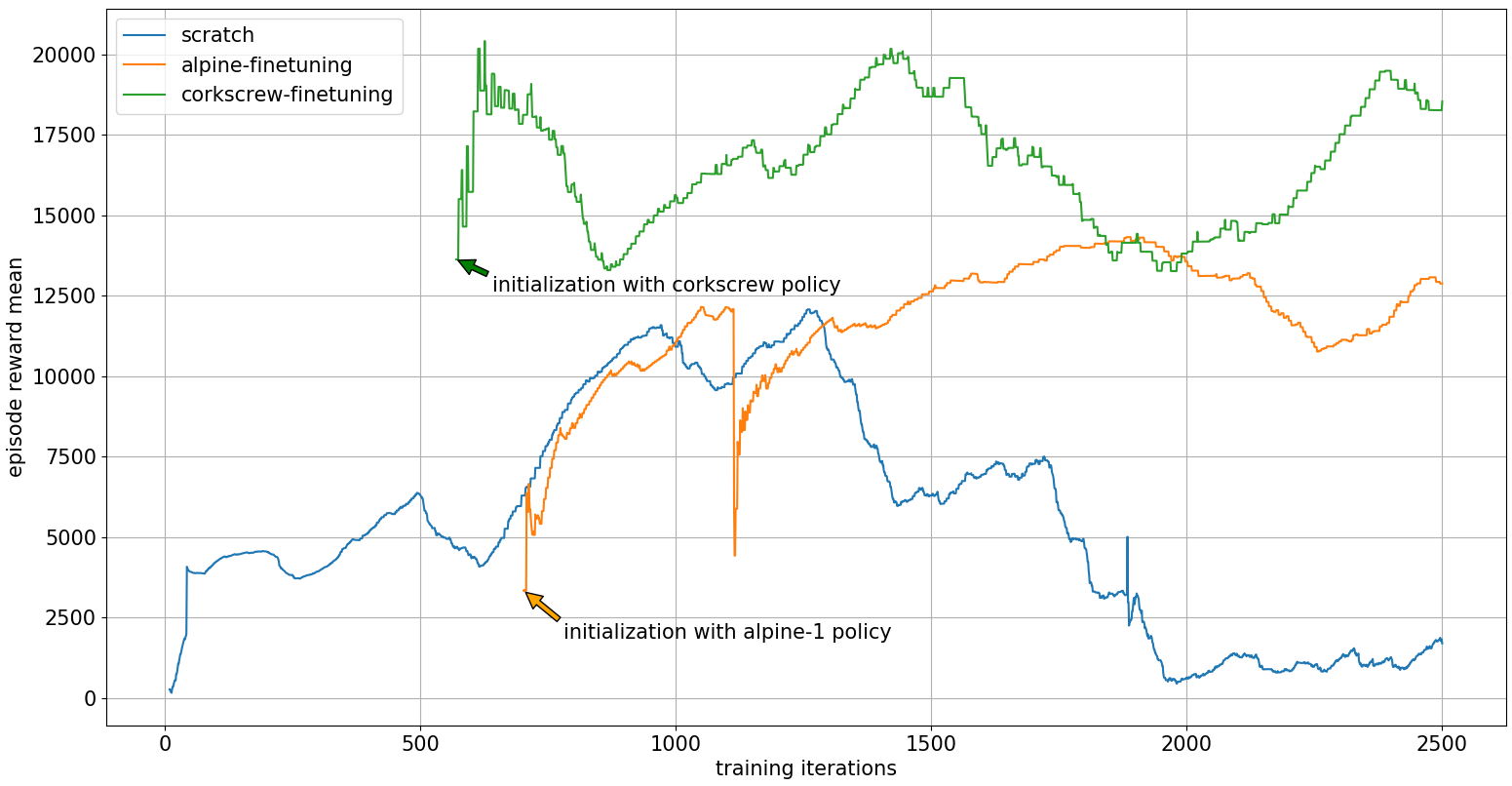}
    \caption{Variation of episode reward over iterations of PPO for learning from scratch on \texttt{spring} compared with first learning on simpler tracks -- \texttt{alpine-1} and \texttt{corkscrew} -- and then fine-tuning on \texttt{spring} (Experiment 3).}
    \label{fig:curriculum_spring}
\end{figure*}

In our third experiment, we present a study to demonstrate how the ability of MADRaS to control the complexity of a driving task in well defined steps can be used to design curricula for an RL agent to accomplish complex tasks in a sample efficient way. We attempt to train a PPO agent to drive \texttt{car1-stock1} on \texttt{Spring} track using the primitive S-A-B action space. With a length of $22.1$ km, \texttt{Spring} is the longest track in TORCS. It has the largest number of turns with different grades of sharpness, both in the left and right directions. It also has ramps and declines. The surface texture varies from place to place. These make it the toughest road track to drive in TORCS. We set the target speed to $100$ Km/hr and maximum episode length to $40000$ steps. Figure \ref{fig:curriculum_spring} and Table \ref{table:curriculum-spring} show the results of this study.  We see that training from scratch on \texttt{Spring} fails to complete one lap of the track even after $2500$ iterations. When we use a curriculum of first training on \texttt{Alpine-1} or \texttt{Corkscrew} tracks followed by fine-tuning on \texttt{Spring} the agent learns to complete the entire lap with high success rates and average speed. In our curriculum learning experiments, we pick the policy that gives the highest mean trajectory reward in the first phase of training (obtained after $701$ iterations in \texttt{Alpine-1} and $561$ iterations in \texttt{Corkscrew}) and use it to initialize the policy in the second phase. The total number of training iterations and the total number of training samples for the curriculum learning strategies (considering both the pre-training and fine-tuning stages) are kept equal to that of training from scratch for fairness of comparison. For fine-tuning, we choose a learning rate of $1\times 10^{-6}$ for the \texttt{Alpine-1} policy and $5\times 10^{-7}$ for the \texttt{Corkscrew} policy. We evaluate the agents in terms of the average fraction of lap covered in an episode, average speed and successful lap completion rate.

\begin{table}[!t]
\centering
\caption{Curriculum learning results for driving in \texttt{Spring} track (Experiment 3).}
\label{table:curriculum-spring}
\begin{tabular}{l|p{1in}|p{1.2in}|p{1in}|}
\cline{2-4}
 & \textbf{Fraction of lap covered} & \textbf{Average Speed (km/hr)} & \textbf{Lap completion rate ($\%$)} \\ \hline
\multicolumn{1}{|l|}{\textbf{Training from scratch}} & \multicolumn{1}{|c|}{0.18} & \multicolumn{1}{|c|}{101.9} & \multicolumn{1}{|c|}{0.0} \\ \hline
\multicolumn{1}{|l|}{\textbf{Pre-training in Alpine-1}} & \multicolumn{1}{|c|}{0.57} & \multicolumn{1}{|c|}{103.5} & \multicolumn{1}{|c|}{27.0} \\ \hline
\multicolumn{1}{|l|}{\textbf{Pre-training in Corkscrew}} & \multicolumn{1}{|c|}{0.54} & \multicolumn{1}{|c|}{100.6} & \multicolumn{1}{|c|}{45.8} \\ \hline
\end{tabular}
\end{table}

\subsection*{Experiment 4: Learning under partial observability and stochastic outcomes of actions}
\label{sec:noise}

\begin{figure*}[!h]
    \centering
    \includegraphics[width=\textwidth]{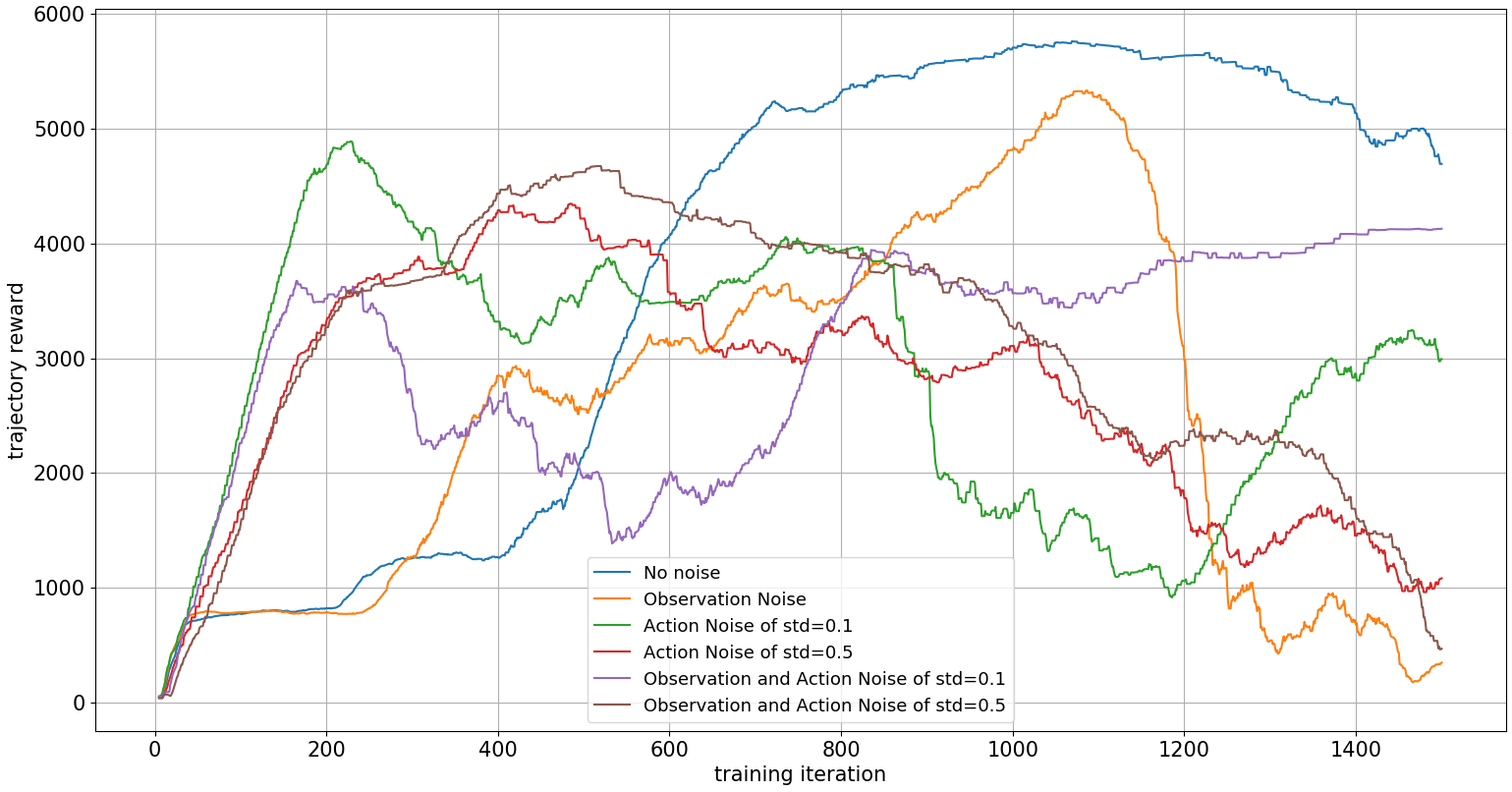}
    \caption{Learning to drive with under partial observability and stochastic outcomes of actions in \texttt{Corkscrew} track (Experiment 4).}
    \label{fig:noise}
\end{figure*}

In this experiment we compare the performances of PPO agents trained to drive \texttt{car1-stock1} around the \texttt{Corkscrew} track with and without observation noise under different levels of stochasticity of the outcome of actions. Observed variables, episode termination criteria and evaluation metrics are the same as in Experiment 1. The reward function is the same as in the Experiments 1-3 (see Table \ref{table:experiment-1-training-criteria}) with the weightage for angular acceleration penalty increased to $8$. As described in Section \ref{sec:madras}, stochastic outcomes of actions is implemented by adding zero mean Gaussian noise to the actions. Figure \ref{fig:noise} shows the learning curves. All these agents are tested in the same track \texttt{Corkscrew} in the presence of both observation noise and $0.5$ standard deviation action noise. Table \ref{table:noise} compares the performance statistics. We observe that the agents trained in the presence of both observation and action noise perform better than the others. This demonstrates the ability of MADRaS to serve as a platform for evaluating the resilience of learning agents to observation noise and environmental stochasticity.

\begin{table*}[]
\centering
\caption{Results of a single PPO agent learning to drive in traffic by RL. The agent was trained to drive in the presence of 4 or 5 traffic cars with equal probability (Experiment 5).}
\label{table:traffic}
\begin{tabular}{l|c|c|c|c|c|c|c|}
\hline
\multicolumn{1}{|p{1.5in}|}{\textbf{Number of traffic agents}} & \textbf{3} & \textbf{4} & \textbf{5} & \textbf{6} & \textbf{7} & \textbf{8} & \textbf{9} \\ \hline
\multicolumn{1}{|p{1.5in}|}{\textbf{Successful task completion rate}} & 99.5\% & 98.1\% & 95.5\% & 96\% & 95.5\% & 95.7\% & 92.8\% \\ \hline
\end{tabular}
\end{table*}

\begin{table}[]
\caption{Learning to drive in the \texttt{corkscrew} track with and without observation noise and different levels of stochasticity in the outcome of actions and evaluation with observation noise and 0.5 std action noise (Experiment 4).}
\label{table:noise}
\centering
\begin{tabular}{l|p{1.5in}|p{1.5in}|}
\cline{2-3}
 & \textbf{Avg. Fraction of Lap Covered} & \textbf{Avg. Speed (km/hr)} \\ \hline
\multicolumn{1}{|p{1.5in}|}{\textbf{No noise}} & \multicolumn{1}{c|}{0.38} & \multicolumn{1}{c|}{52.54} \\ \hline
\multicolumn{1}{|p{1.5in}|}{\textbf{Observation noise}} & \multicolumn{1}{c|}{0.19} & \multicolumn{1}{c|}{30.78} \\ \hline
\multicolumn{1}{|p{1.5in}|}{\textbf{Stochastic actions (noise std 0.1)}} & \multicolumn{1}{c|}{0.12} & \multicolumn{1}{c|}{29.99} \\ \hline
\multicolumn{1}{|p{1.5in}|}{\textbf{Stochastic actions (noise std 0.5)}} & \multicolumn{1}{c|}{0.64} & \multicolumn{1}{c|}{48.67} \\ \hline
\multicolumn{1}{|p{1.5in}|}{\textbf{Observation noise and Stochastic actions (noise std 0.1)}} & \multicolumn{1}{c|}{0.63} & \multicolumn{1}{c|}{48.85} \\ \hline
\multicolumn{1}{|p{1.5in}|}{\textbf{Observation noise and Stochastic actions (noise std 0.5)}} & \multicolumn{1}{c|}{0.68} & \multicolumn{1}{c|}{46.91} \\ \hline
\end{tabular}
\end{table}

\pdfoutput=1
\subsection*{Experiment 5: Learning to drive in traffic}
\label{sec:traffic}

\begin{figure}
    \centering
    \includegraphics[width=0.8\textwidth]{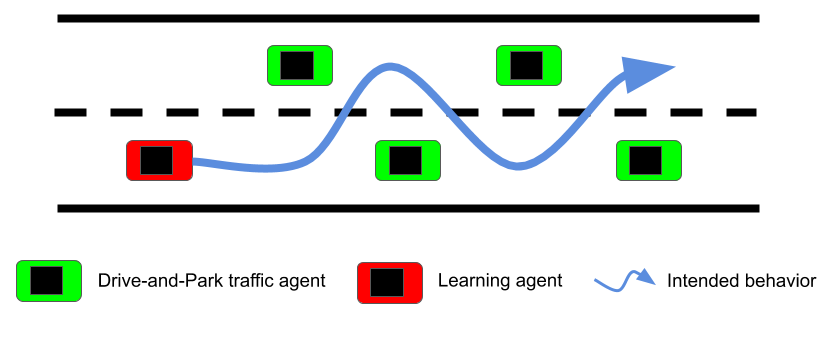}
    \caption{Schematic diagram of the environment design for Experiment 5. The task of the learning agent is to overtake all the traffic cars without colliding with any of them or going off track.}
    \label{fig:training-in-traffic}
\end{figure}

\begin{table}[!t]
\centering
\caption{RL training criteria for Experiment 5. Please refer to \cite{loiacono2013simulated} for details on the observed variables.}
\label{table:experiment-traffic-training-criteria}
\begin{tabular}{|l|l|c|}
\hline
\multirow{6}{*}{\textbf{Reward function}} & \textbf{Reward Function Component} & \textbf{Weightage} \\ \cline{2-3} 
 & Progress Reward & 1.0 \\ \cline{2-3} 
 & Average Speed Reward & 1.0 \\ \cline{2-3} 
 & Collision Penalty & 10.0 \\ \cline{2-3} 
 & Turn Backward Penalty & 10.0 \\ \cline{2-3} 
 & Angular Acceleration Penalty & 1.0 \\ \cline{2-3}
 & Overtake Reward & 5.0 \\ \cline{2-3}
 & Rank 1 Reward & 100.0 \\ \hline\hline
\textbf{Observed variables} & \multicolumn{2}{p{3in}|}{angle, track, trackPos, speedX, speedY, speedZ, opponents} \\ \hline\hline
\textbf{Done criteria} & \multicolumn{2}{p{3in}|}{Rank 1, Time Out, Collision, Turn Backward, Out of Track} \\ \hline
\end{tabular}
\end{table}

In this experiment we use the ability of MADRaS to generate custom traffic to train an agent to navigate through a narrow road without colliding with any traffic car -- moving or parked. Figure \ref{fig:training-in-traffic} shows a schematic diagram of the training environment. We choose the \texttt{Aalborg} track for this study since it is one of the narrowest tracks of TORCS and further reduce its width to half resulting in an effective track width of $5$m.\\ 

The traffic agents used in this experiment are \texttt{DriveAndParkAgents} (see Table \ref{table:traffic-agents}). MADRaS positions the traffic cars ahead of the learning car at the start of the race. When an episode begins, the \texttt{DriveAndParkAgents} start driving at their given target speeds $(50)$ km/hr towards their given parking locations (specified in terms of distance from the start of the race and track position) using PID controllers. This way, the learning agent sees moving cars in the beginning and parked cars towards the end of each episode. This forces it to learn to avoid collision with both static and moving obstacles. We set the parking locations of the traffic cars on alternate sides of the road so that the the agent must learn to turn both left and right to overtake all the traffic cars. We maintain a gap of at least $10$m between consecutive parking locations along the length of the road to make sure that the learning car has enough space to maneuver between the traffic cars. To create variance in the environment, we randomly vary each parking location within an area of $5$m along the track length and $0.25$m along the track width. We also switch the number of traffic cars between $4$ and $5$ with equal probability. Changing the number of traffic cars also makes sure that the learning agent gets initialized in the left and right halves of the track with equal probability. We use the T-S action space and set the target speed of the learning agent to $50$ km/h. Table \ref{table:experiment-traffic-training-criteria} gives the training criteria for this experiment. \\

The agent gets an \emph{Overtake Reward} every time it overtakes a traffic agent and \emph{Rank 1 Reward} at the end of the episode if it manages to overtake all the traffic agents. The agent is evaluated in terms of the fraction of times it overtakes all the traffic cars successfully.\\

Table \ref{table:traffic} presents the results of this experiment. We observe that the agent learns to generalize to both fewer and more traffic agents than it encountered during training and navigate its way through them collision-free with a high success rate.\\

\subsection*{Experiment 6: Multi-agent reinforcement learning}
\label{sec:multi-agent}
\begin{figure}
    \centering
    \includegraphics[width=0.8\textwidth]{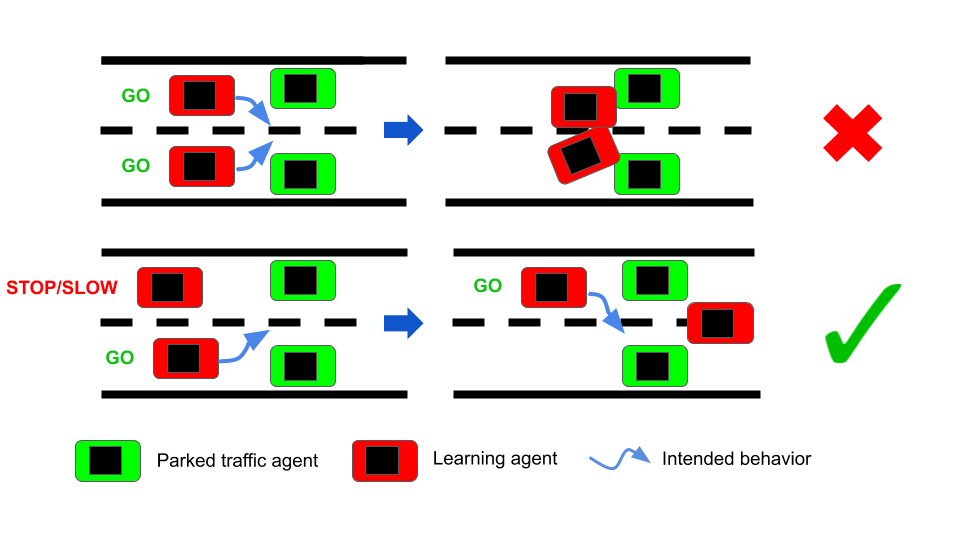}
    \caption{Schematic diagram of the multi-agent task studied in Experiment 6. The task for the two learning agents is to coordinate with each other and pass through the gap between the parked traffic cars without making any collision. The top row shows an example of undesirable behavior in which both the agents attempt to pass through the bottleneck at the same time and result in a collision. The bottom row gives a feasible solution to the problem in which one of the agents stops or slows down to wait for the other agent to pass through the gap. Only after the gap is clear does it attempt to pass through -- thus avoiding collision with any of the other cars.}
    \label{fig:multi-agent-schematic}
\end{figure}

\begin{figure}[]
    \centering
    \subfloat[Agent-1 Training Curves]{{\includegraphics[width=0.7\textwidth]{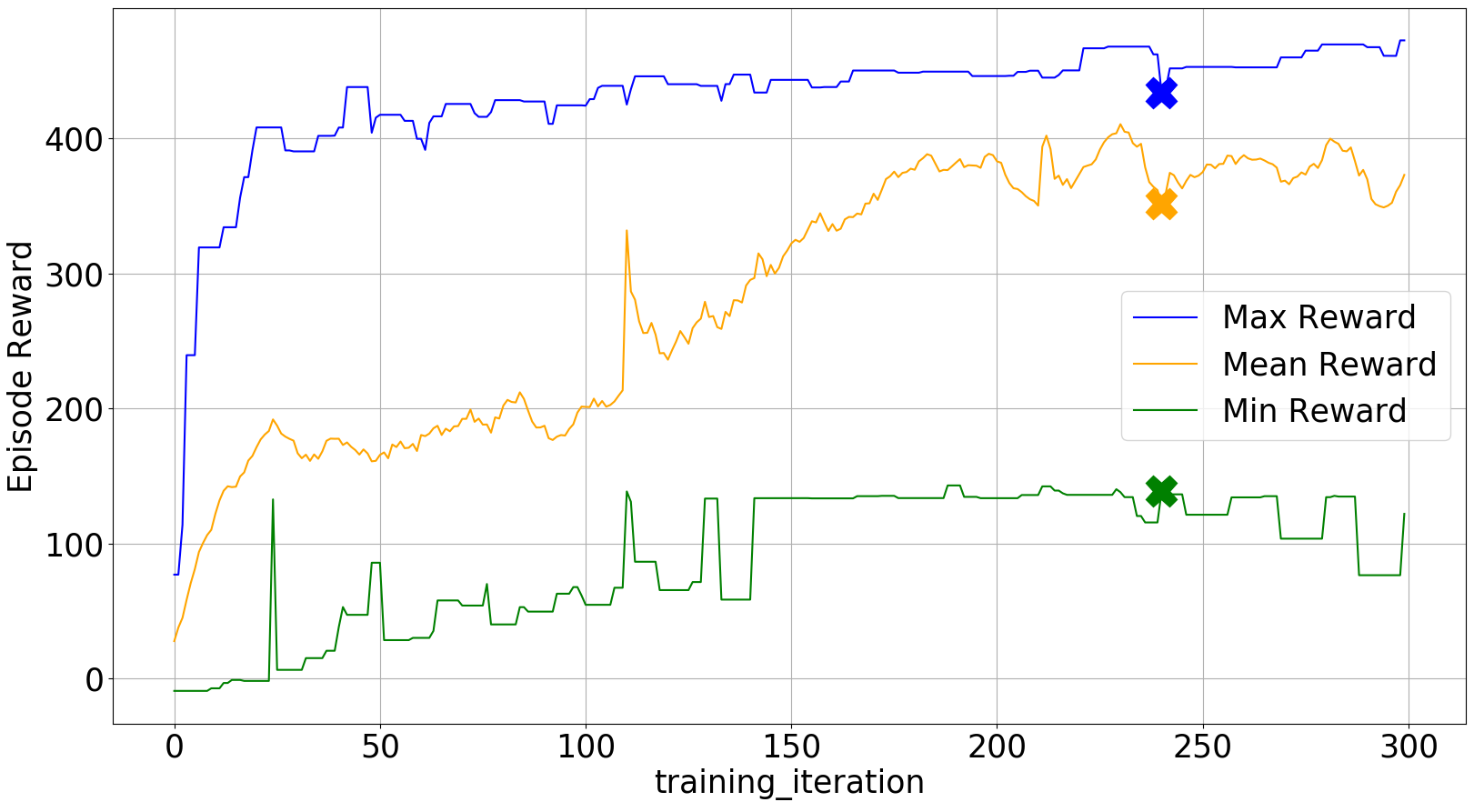}}}\\%
    \subfloat[Agent-2 Training Curves]{{\includegraphics[width=0.7\textwidth]{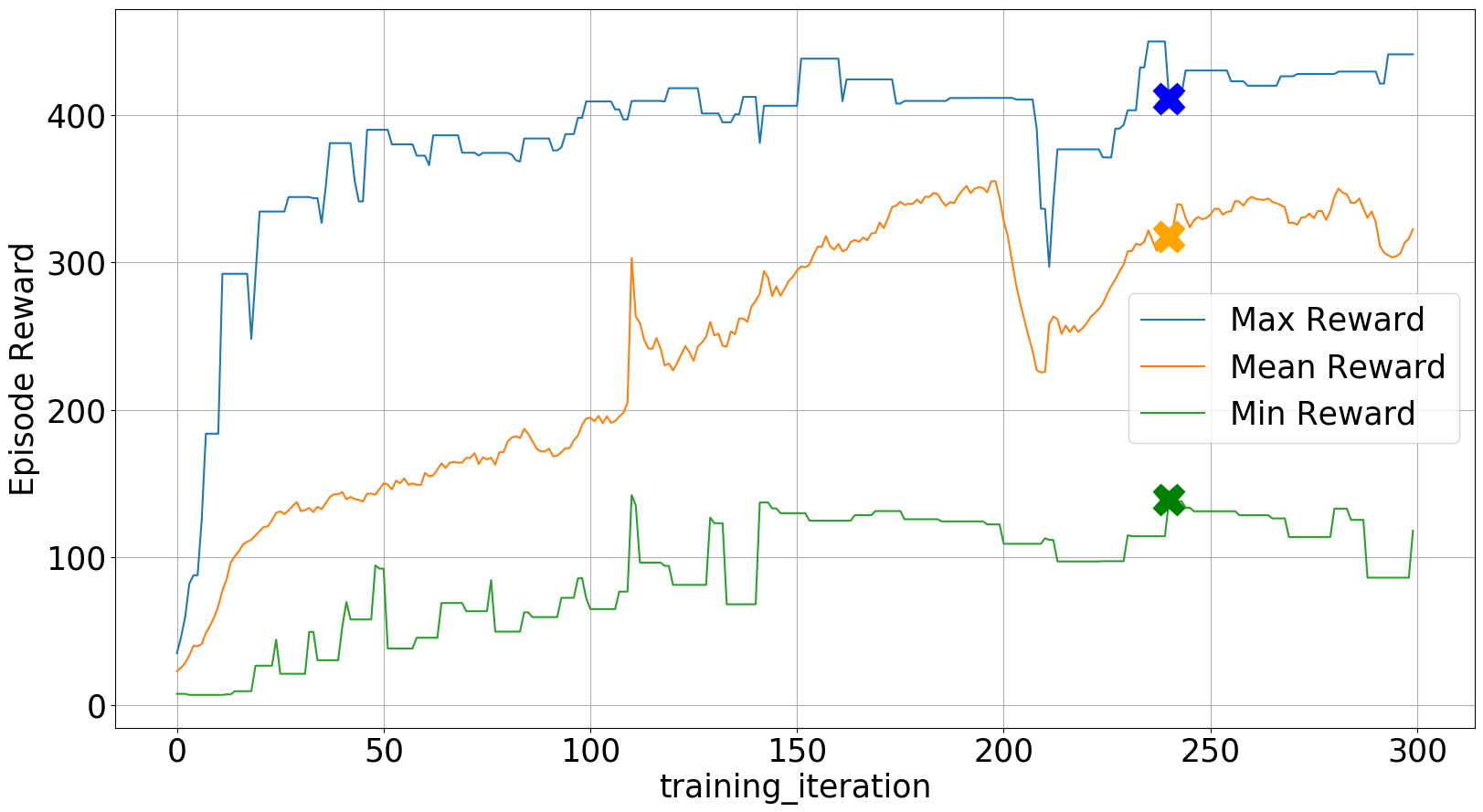}}}\\%
    \subfloat[Joint learning Curves]{{\includegraphics[width=0.7\textwidth]{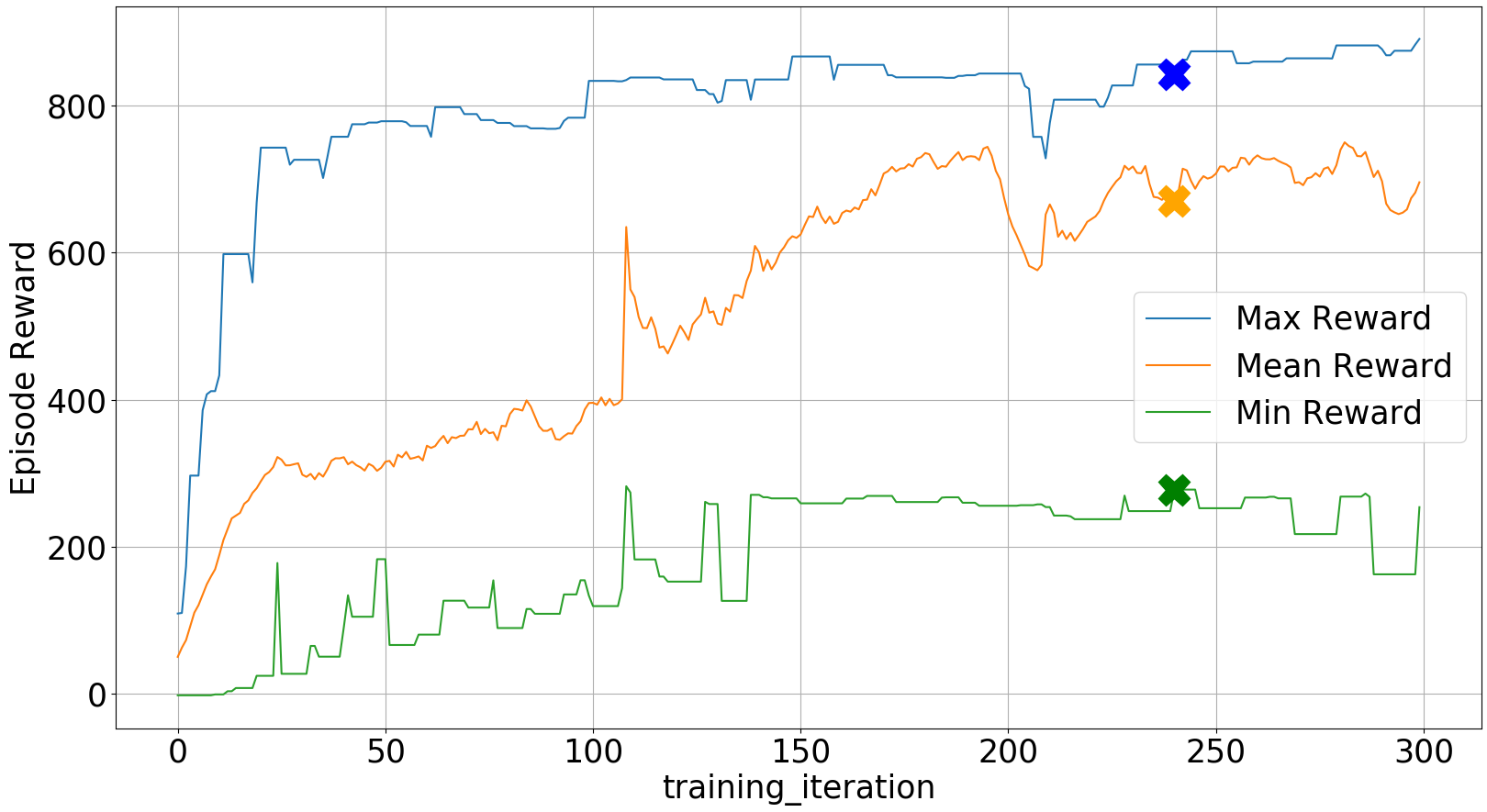}}}\\%
    \caption{Learning curves for multi-agent training in Experiment 6. The cross symbol denotes transition point in the agent's curriculum where the first task ends and the second task begins.}
    \label{fig:agents}
\end{figure} 

\begin{table}[!t]
\centering
\caption{Dimensions of cars used in Experiment 6.}
\label{table:cars-used-in-marl-experiment}
\begin{tabular}{l|c|c|c|}
\cline{2-4}
 & \textbf{Car Model} & \textbf{Length (m)} & \textbf{Width (m)} \\ \hline
\multicolumn{1}{|l|}{\textbf{Traffic Car}} & \texttt{car1-trb1} & 4.52 & 1.94 \\ \hline
\multicolumn{1}{|l|}{\textbf{PPO Agent-1}} & \texttt{car3-trb1} & 4.55 & 1.95 \\ \hline
\multicolumn{1}{|l|}{\textbf{PPO Agent-2}} & \texttt{car5-trb1} & 4.67 & 1.94 \\ \hline
\end{tabular}
\end{table}

\begin{table}[]
\centering
\caption{Curriculum for multi-agent RL in Experiment 6.}
\label{table:multi-agent-curriculum}
\begin{tabular}{|c|c|c|}
\hline
\textbf{Iterations of training} & \textbf{Parking Distance (m)} & \textbf{Gap Width (m)} \\ \hline
1--240 & 30--40 & 2.76--4.06 \\ \hline
240--300 & 30--35 & 2.76--3.46 \\ \hline
\end{tabular}
\end{table}

\begin{table}[!t]
\centering
\caption{RL training criteria for Experiment 6. \emph{peerActions} refers to the actions of the other learning agent from the previous time step. Please refer to \cite{loiacono2013simulated} for details on the other observed variables.}
\label{table:experiment-3-training-criteria}
\begin{tabular}{|l|l|c|}
\hline
\multirow{6}{*}{\textbf{Reward function}} & \textbf{Reward Function Component} & \textbf{Weightage} \\ \cline{2-3} 
 & Progress Reward & 1.0 \\ \cline{2-3} 
 & Average Speed Reward & 1.0 \\ \cline{2-3} 
 & Collision Penalty & 10.0 \\ \cline{2-3} 
 & Turn Backward Penalty & 10.0 \\ \cline{2-3} 
 & Angular Acceleration Penalty & 5.0 \\ \hline\hline
\textbf{Observed variables} & \multicolumn{2}{p{3in}|}{angle, track, trackPos, opponents, speedX, speedY, speedZ, peerActions} \\ \hline\hline
\textbf{Done criteria} & \multicolumn{2}{p{3in}|}{Race Over, Time Out, Collision, Turn Backward, Out of Track} \\ \hline
\end{tabular}
\end{table}

One of the biggest aspirations of autonomous driving is the avoidance of traffic congestion through cooperation. In this experiment we utilize the multi-agent training ability of MADRaS and its framework for inter-vehicular communication to solve a simplified version of this task by multi-agent reinforcement learning.\\

The training environment consists of two PPO agents and two traffic agents on the \texttt{Corkscrew} track. The PPO agents communicate their actions to each other at every step. We park the traffic agents next to each other with a small gap in between that is sufficient only for one car to pass through. The task of the PPO agents is to pass through the gap one by one without colliding with each other or with any traffic agent (see Figure \ref{fig:multi-agent-schematic}). Thus the agents must learn a collaborative strategy in which the agent trying to pass through the gap first should be given enough time to pass through completely by the other agent before it makes its own attempt.\\

Table \ref{table:cars-used-in-marl-experiment} gives the cars assigned to the learning and traffic agents and their dimensions. Both the PPO agents have T-S action space. Table \ref{table:multi-agent-curriculum} describes the curriculum used for the training. We randomly vary the parking distance of each traffic car and the gap between them at the start of each episode for improved generalization. Table \ref{table:experiment-3-training-criteria} gives the details of the observed variables, reward functions and done criteria. The agents must learn the following distinct skills to be able to accomplish this task.
\begin{itemize}
    \item Running forward without going off track.
    \item Not colliding with each other.
    \item Not colliding with any of the parked cars.
    \item Learning to collaborate and pass through the bottleneck one by one.
\end{itemize}

We jointly evaluate the agents in terms of the rate of successful passage of both the agents through the traffic bottleneck. Figure \ref{fig:agents} shows the individual and joint learning curves respectively during training. The final evaluation is done over $100$ episodes of stochastically parked agents and the PPO agents demonstrate a joint task completion rate of \textbf{$83.3\%$}.

\pdfoutput=1
\section{Conclusion}
\label{sec:conclusion}
In this paper we present MADRaS, an open-source Multi-Agent Driving Simulator for autonomous driving. MADRaS builds on TORCS, a popular car racing platform, and adds a suite of features like hierarchical control modes, domain randomization, custom traffic, partial observability, stochastic outcomes of actions and support for multi-agent training. We train reinforcement learning agents to accomplish challenging tasks like generalizing across a wide range of track geometries and vehicular dynamics, driving under stochasticity and partial observability, navigating through static and moving obstacles and negotiating with other agents to pass through a traffic bottleneck. These studies demonstrate the viability of MADRaS to simulate rich highway and track driving scenarios of high variance and complexity that are valuable for autonomous driving research. We wish to develop features specific to fuel management and vehicular safety in the future.
\pdfoutput=1
\section*{Acknowledgement}
The authors would like to thank Professor Pabitra Mitra of the Department of Computer Science and Engineering, IIT Kharagpur for his helpful feedback on the structure of the paper and Manish Prajapat of ETH Zurich for his useful tips on the implementation of inter-vehicular communication in MADRaS. The authors would also like to thank Intel Labs India for incubating the early stage of this project. Anirban Santara's work in this project was supported by Google India, under the Google India PhD Fellowship grant, and Intel Inc. under the Intel Student Ambassador Program.

\newpage
\vskip 0.2in
\bibliography{references}
\bibliographystyle{theapa}

\newpage
\appendix
\setcounter{table}{0}
\renewcommand{\thetable}{A\arabic{table}}
\begin{appendices}
\section*{Appendix A. Configuring MADRaS}
\label{Appendix:A}
The structure of MADRaS focuses on ease of use and encourages custom modifications. In this section we describe the configuration variables of MADRaS. All these variables are specified in the \texttt{envs/data/madras\_config.yml} file. The `yaml' (or `yml') format provides a powerful yet convenient way of specifying most data types and basic data structures like lists and dictionaries.\\

\noindent The \texttt{madras\_config.yml} file has three sections:
\begin{enumerate}
    \item \textbf{Server configuration:} In this section the user can set the global configuration of the MADRaS environment. Since MADRaS can randomly vary the driving tracks, model of car for the learning agents, and the number of traffic cars between episodes, these terms are specified as lists and ranges. The maximum total number of cars (including learning and traffic agents) in the environment can be specified as \texttt{max\_cars} and the minimum number of traffic cars by \texttt{min\_traffic\_cars}. The number of learning agents ($N_{l}$) is specified in the ``agent configuration" section. $$N_l + \text{\texttt{min\_traffic\_cars}} \le \text{\texttt{max\_cars}}$$ The environment becomes deterministic with respect to the number of traffic cars if equality holds. Otherwise the number of traffic cars on the track is randomly chosen between \texttt{min\_traffic\_cars} and $\text{\texttt{max\_cars}}-N_l$. The list of car models to choose randomly for the learning agent can be specified in \texttt{learning\_car}. The list of tracks to choose randomly for each episode can be specified in \texttt{track\_names}.

    \item \textbf{Agent configuration:} In the \texttt{agents} section, the user can specify the configurations of the learning agents. They can set the \texttt{target\_speed}, \texttt{pid\_settings} for the low level controller if \texttt{pid\_assist} is \texttt{True}, configuration of the observation space (according to the modes in \texttt{utils/observation\_handler.py}), reward function (to be parsed by \texttt{utils/reward\_handler.py}) and done function (to be parsed by \texttt{utils/done\_handler.py}).

    \item \textbf{Traffic configuration:} In the \texttt{traffic} section, the user can specify the details of the traffic agents to be used in the simulation. If $N_t$ traffic agents need to be chosen in a given episode, their configurations will be set to the first $N_t$ elements from the list of agents in this section. These configurations are parsed by \texttt{traffic/traffic.py}. The \texttt{target\_speed}, \texttt{target\_lane\_pos}, collision avoidance properties and \texttt{pid\_settings} of the traffic cars can be specified here. If the traffic agents need to be parked in certain locations (specified in terms of their distance from the start line and lane position) of the track before the start of each episode, that can also be specified in this section. 
\end{enumerate}

\noindent The full list of the configuration variables is available in Tables \ref{table:server-config}, \ref{table:agent-config} and \ref{table:traffic-config}.\\

\noindent MADRaS supports inter-vehicular communication (IV-Comm) between the learning agents. The settings for the IV-Comm system can be specified in \texttt{envs/data/communications.yml}. The user can specify the list of variables (\texttt{vars}) that each learning agent wants to observe from a list of communicating agents (\texttt{comms}) for a given number of previous time steps (\texttt{buff\_size}).

\begin{table*}[b]
\centering
\caption{Server Configuration Parameters}
\label{table:server-config}
\begin{tabular}{|l|l|l|}
\hline
\textbf{Parameters}        & \textbf{Description}                                                                                                                           & \textbf{Possible Values}                                                  \\ \hline
torcs\_server\_port        & \begin{tabular}[c]{@{}l@{}}For setting the port of communication\\ with the TORCS Server.\end{tabular}                                         & $\mathbb{Z}^{+}$                              \\ \hline
max\_cars                  & Max number of vehicles to be spawned.                                                                                                          & $\mathbb{Z}^{+}$                              \\ \hline
track\_names               & \begin{tabular}[c]{@{}l@{}}List of tracks on which the simulation\\ will run.\end{tabular}                                                     & \begin{tabular}[c]{@{}l@{}}List of track \\ names\end{tabular}            \\ \hline
track\_limits                  & \begin{tabular}[c]{@{}l@{}} Restrict the agent to remain within a given\\ range of track\_pos values.\end{tabular}                                                                                                          & $(\mathbb{R}, \mathbb{R})$                              \\ \hline
distance\_to\_start        & \begin{tabular}[c]{@{}l@{}}Starting distance of the cars from the \\ start line.\end{tabular}                                                  & $\mathbb{Z}^{+}$                             \\ \hline
torcs\_server\_config\_dir & \begin{tabular}[c]{@{}l@{}}The location of the TORCS server racing \\ config directory.\end{tabular}                                           & Path string                                                               \\ \hline
scr\_server\_config\_dir   & The location of available cars config directory                                                                                                & Path string                                                               \\ \hline
traffic\_car               & The type of car to be used for traffic                                                                                                         & car name                                                                  \\ \hline
learning\_car              & \begin{tabular}[c]{@{}l@{}}List of car models for using as the learning \\ agent.\end{tabular}                                                 & \begin{tabular}[c]{@{}l@{}}List of car\\ names\end{tabular}               \\ \hline
randomize\_env             & \begin{tabular}[c]{@{}l@{}}Flag for setting randomization on. (Cycles through\\ the selected cars and tracks in a random fashion)\end{tabular} & boolean                                                                   \\ \hline
add\_noise\_to\_actions    & \begin{tabular}[c]{@{}l@{}}Flag for adding a small Gaussian Noise to the \\ actions before sending to the TORCS server.\end{tabular}           & boolean                                                                   \\ \hline
action\_noise\_std         & \begin{tabular}[c]{@{}l@{}}Specifies the standard deviation of the Gaussian \\ for the noise addition.\end{tabular}                            & $[0, 1]$                                                                \\ \hline
noisy\_observations        & \begin{tabular}[c]{@{}l@{}}Toggles the TORCS flag for enabling noisy\\ observations.\end{tabular}                                              & boolean                                                                   \\ \hline
visualise                  & Flag for setting the display on and off.                                                                                                       & boolean                                                                   \\ \hline
no\_of\_visualisations     & To visualize multiple training instances                                                                                                       & $\mathbb{Z}^{+}$                              \\ \hline
max\_steps                 & \begin{tabular}[c]{@{}l@{}}Maximum steps that the environment will take \\ before resetting.\end{tabular}                                      & $\mathbb{Z}^{+}$                              \\ \hline
\end{tabular}
\end{table*}


\pdfoutput=1
\begin{table*}[]
\centering
\caption{Agent Configuration Parameters}
\label{table:agent-config}
\begin{tabular}{|l|l|l|}
\hline
\textbf{Parameters}                                                                                & \textbf{Description}                                                                                              & \textbf{Possible Values}                                         \\ \hline
vision                                                                                             & \begin{tabular}[c]{@{}l@{}}Flag for activating visual input instead of the usual\\ sensor based one.\end{tabular} & boolean                                                          \\ \hline
throttle                                                                                           & Flag for activating throttle control on and off.                                                                  & boolean                                                          \\ \hline
gear\_change                                                                                       & Flag for activating gear control on and off.                                                                      & boolean                                                          \\ \hline
client\_max\_steps                                                                                 & Maximum steps that the client is available to take.                                                               & $\mathbb{Z}^{+} \cup \{-1\}$                                                             \\ \hline
target\_speed                                                                                      & Target speed setting of the agent car.                                                                            & $\mathbb{Z}^{+}$                                                                \\ \hline
state\_dim                                                                                         & Dimension of the Observation Space.                                                                               & $\mathbb{Z}^{+}$                                                                \\ \hline
normalize\_actions                                                                                 & Toggle to turn on action normalization.                                                                            & boolean                                                          \\ \hline
pid\_assist                                                                                        & Toggle to turn on T-S control mode.                                                                               & boolean                                                          \\ \hline
pid\_settings{[}accel\_pid{]}                                                                      & $K_p$, $K_i$, $K_d$ for throttle PID.                                                                                      & List of floats                                                   \\ \hline
pid\_settings{[}accel\_pid{]}                                                                      & $K_p$, $K_i$, $K_d$ for steering PID.                                                                                      & List of floats                                                   \\ \hline
accel\_scale                                                                                       & Acceleration Scaling.                                                                                             & $\mathbb{R}^{+}$                                                                \\ \hline
steer\_scale                                                                                       & Steering Scaling.                                                                                                 & $\mathbb{R}^{+}$                                                                \\ \hline
pid\_latency                                                                                       & \begin{tabular}[c]{@{}l@{}}Number time-steps the control command sticks \\ to the server.\end{tabular}            & $\mathbb{Z}^{+}$                                                                \\ \hline
observations{[}mode{]}                                                                             & Name of the Observation Class.                                                                                    & string                                                           \\ \hline
\textit{\begin{tabular}[c]{@{}l@{}}observations{[}multi\_flag{]}\\ (multi mode only)\end{tabular}} & \begin{tabular}[c]{@{}l@{}}Toggle for turning on communication for the\\ agent i,\end{tabular}                    & boolean                                                          \\ \hline
observations{[}buff\_size{]}                                                                       & Specifies the buffer size of action.                                                                              & $\mathbb{Z}^{+}$                                                                \\ \hline
observation{[}normalize{]}                                                                         & Toggle to tun on observation normalization.                                                                       & boolean                                                          \\ \hline
obs\_min                                                                                           & Minimum values for certain observation attributes.                                                                & dict                                                             \\ \hline
obs\_max                                                                                           & Maximum values for certain observation attributes.                                                                & dict                                                             \\ \hline
rewards{[}name, scale{]}                                                                           & \begin{tabular}[c]{@{}l@{}}List of the Reward classes and a scaling factor of \\ the rewards.\end{tabular}        & \begin{tabular}[c]{@{}l@{}}list of names\\ and dict\end{tabular} \\ \hline
dones                                                                                              & Done conditions currently in use.                                                                                 & list of dones                                                    \\ \hline
\end{tabular}
\end{table*}





\begin{table*}[]
\centering
\caption{Common Traffic Configuration Parameters}
\label{table:traffic-config}
\begin{tabular}{|l|l|l|}
\hline
\textbf{Parameters}           & \textbf{Description}                                                                           & \textbf{Possible Values} \\ \hline
Name                          & Traffic Agent Type,                                                                            & string      \\ \hline
target\_speed                 & Traffic Agent Speed.                                                                           & $\mathbb{R}^+$                        \\\hline
initial\_distance                 & Traffic Agent initial distance from start line (range).                                                                           & 2-Tuple of Floats                       \\\hline
initial\_trackpos                 & Traffic Agent initial track-position (range).                                                                           & 2-Tuple of Floats                        \\\hline
track\_len                    & Length of the Current Track.                                                                   & $\mathbb{R}^+$                        \\ \hline
pid\_settings{[}accel\_pid{]} & $K_p$, $K_i$, $K_d$ values for acceleration.                                                            & List of Floats                     \\ \hline
pid\_settings{[}steer\_pid{]} & $K_p$, $K_i$, $K_d$ values for steering.                                                                & List of Floats                     \\ \hline
accel\_scale                  & Acceleration scaling.                                                                          & $\mathbb{R}^+$                        \\ \hline
steer\_scale                  & Steering scaling.                                                                              & $\mathbb{R}^+$                        \\ \hline
collision\_time\_window       & \begin{tabular}[c]{@{}l@{}}Describes the collision region\\ for the traffic agent\end{tabular} & $\mathbb{R}^+$                        \\ \hline
\end{tabular}
\end{table*}




\section*{Appendix B. PID Response}
\label{Appendix:B}

\begin{figure}[]
    \centering
    \subfloat[Track Position PID Response Curve]{{\includegraphics[width=\textwidth]{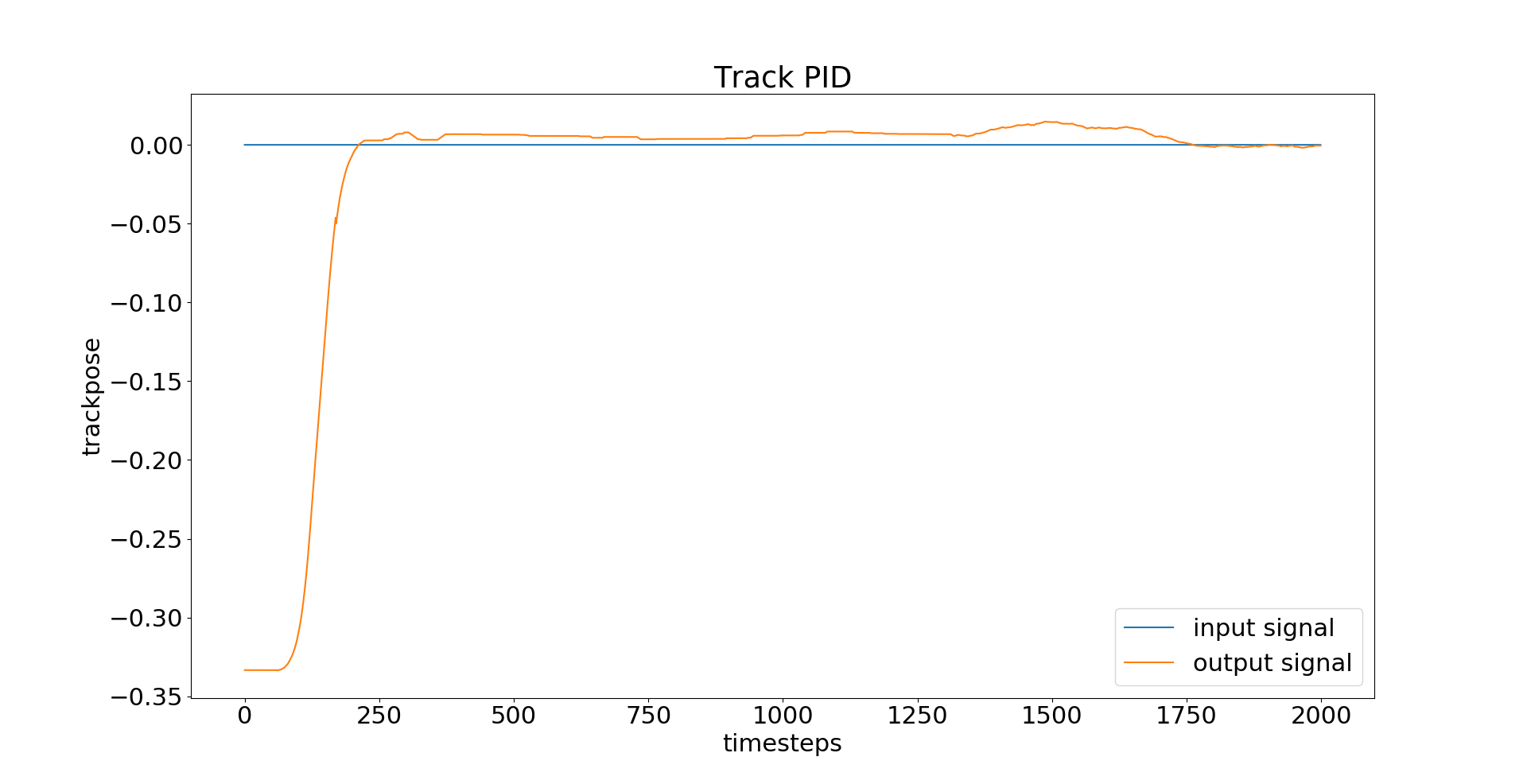}}}\\%
    \subfloat[Speed PID Response Curve]{{\includegraphics[width=\textwidth]{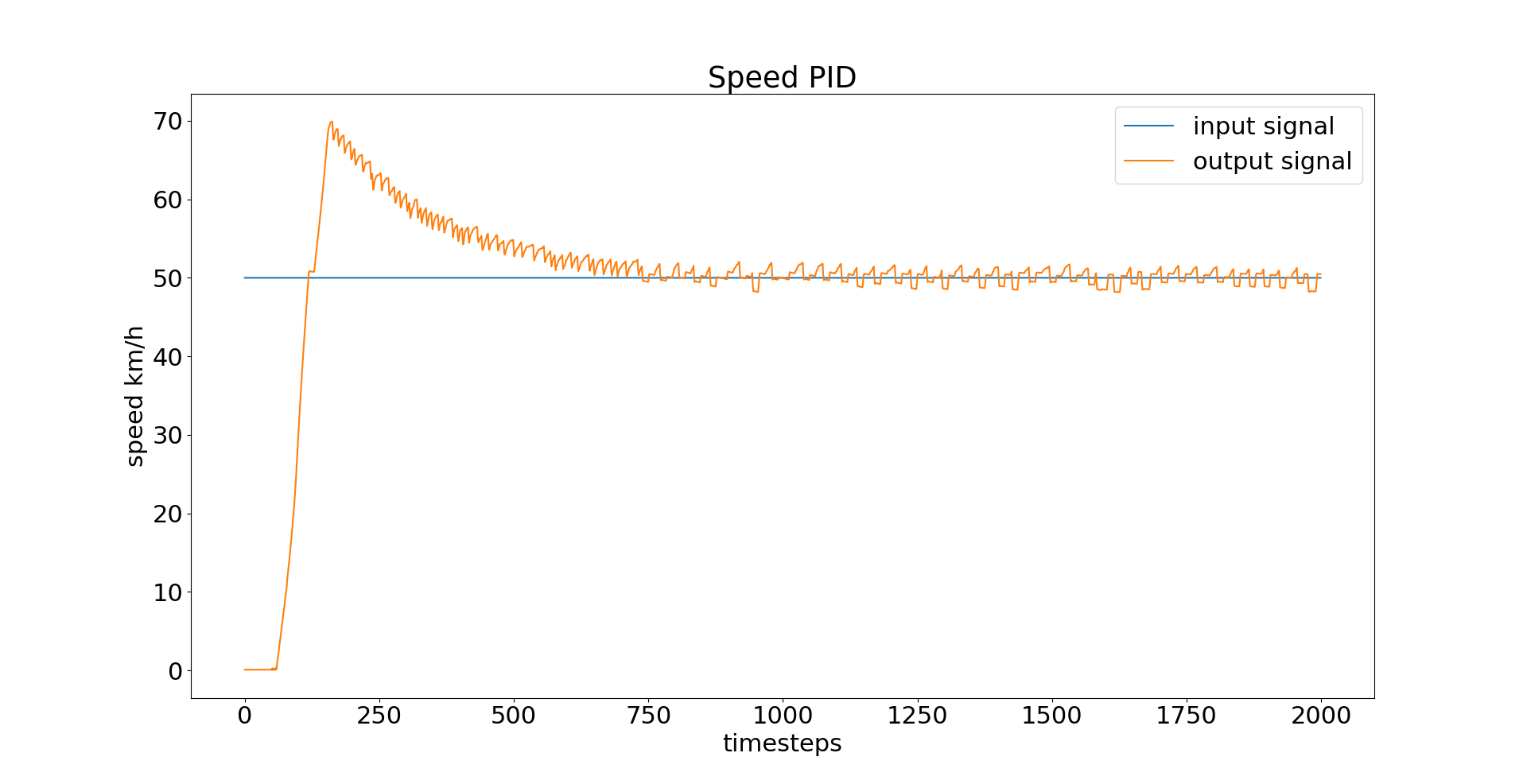}}}\\%
    \caption{Sample response curves of the vanilla PID controllers used for Track-Position and Speed in MADRaS. The \textit{Input Signal} represents the target and the \textit{Output Signal} represents the response that the controller produces.}
    \label{fig:pid}
\end{figure} 

In this section we describe our implementation of the PID controller for the optional high-level track-position -- speed control mode in MADRaS. Please note that this implementation can be easily swapped out for a more sophisticated one by creating a derived class of \texttt{PIDController} defined in \texttt{controllers/pid.py}. The error function for track-position PID controller is defined as a function of the track-pose and the angle that the car's heading makes with the center line. The output of this controller is the steer-angle of the vehicle for the current time-step that would bring the car closer to the desired track-position.
\begin{equation}
    e(t) = \theta(t-1) - (TP(t-1) - TP_{\textit{target}}) * \textit{scale}
\end{equation}

\noindent The error function for the Speed PID controller is a function of the forward velocity. The output of the controller is the value of acceleration and braking that would bring the speed closer to the target speed of the vehicle.

\begin{equation}
    e(t) = (V(t-1) - V_{\textit{target}}) * \textit{scale}
\end{equation}

Where:
\begin{itemize}
    \item $e(t):$ Error at time $t$.
    \item $TP(t):$ Track-Pose at time $t$.
    \item $V(t):$ Forward Velocity at time $t$.
    \item \textit{scale}: Scaling Factor.
\end{itemize}

Figure \ref{fig:pid} presents some sample responses of the PID controllers to a given target track-position and speed.

\section*{Appendix C. Initial State Distribution}
\label{Appendix:C}

MADRaS has a \texttt{randomize\_env} flag where the traffic cars and agents are randomly placed on the race track. As previously mentioned that the configuration of a vehicle in the simulator is based on four different attributes.
\begin{itemize}
    \item Vehicle Model: The model of the car assigned to the learning agent(s) is randomly selected from the already specified list of vehicles from a Categorical distribution.
    \item Number of Traffic Vehicles : The number of vehicles is randomly selected within the specified range using a Discrete Uniform distribution. 
    \item Initial Track Position : The initial track position of each traffic vehicle is selected randomly between the two edges of the road (specified by the \texttt{initial\_trackpos} tuple) from a Continuous Uniform distribution. 
    \item Distance from Start : The distance from start for the learning agent is fixed but for the specified traffic agent it is set between a fixed range (specified by the \texttt{initial\_distance} tuple) from a Continuous Uniform distribution. 
\end{itemize}

\end{appendices}
\end{document}